\documentclass[fleqn,10pt]{wlscirep}

\usepackage[utf8]{inputenc}
\usepackage[T1]{fontenc}
\usepackage{lmodern}
\usepackage{amsmath}
\usepackage{amssymb}
\usepackage{amsfonts}
\usepackage{booktabs}
\usepackage{array}
\usepackage{graphicx}
\usepackage[labelfont=bf]{caption}
\usepackage{subcaption}
\usepackage{hyperref}
\usepackage{float}
\usepackage{xcolor}

\usepackage{multirow}
\usepackage{longtable}

\title{Response drift across frontier large language models}

\author[1,*]{Mohammed Aledhari}
\author[1,+]{Ali Aledhari}
\author[1,+]{Fatimah Aledhari}
\author[1,+]{Gowtham Venkat Eathamokkala}
\author[2,+]{Mohamed Rahouti}
\affil[1]{Department of Data Science, University of North Texas, Denton, TX 76207, USA}
\affil[2]{Department of Computer and Information Science, Fordham University, New York, NY 10458, USA}
\affil[*]{mohammed.aledhari@unt.edu}
\affil[+]{these authors contributed equally to this work}

\begin{abstract}
All frontier large language models (LLMs) exhibit response drift---producing 
outputs that deviate from expert-validated references---yet the magnitude 
and structure of this drift remain uncharacterised by systematic human 
evaluation. Here we report a fully crossed evaluation in which 
47~geographically diverse participants each assessed all 62~multidomain 
questions across ten frontier LLMs under blinded conditions, yielding 
29,140 independent assessments. Every model drifts, but drift magnitude 
varies substantially: eight models converge on a statistically 
indistinguishable ceiling (78--81\% deviation), while two achieve lower 
deviation (47--49\%). Drift profiles differ across six domains and 62 
questions, with pairwise correlations among ceiling models exceeding 
$r = 0.85$. Automated similarity metrics explain less than 2\% of 
variance in human judgements. These findings reveal that response drift 
is universal across frontier LLMs, domain- and question-dependent in 
structure, and accessible only through human-centred evaluation.
\end{abstract}

\definecolor{boxbg}{HTML}{f5f7fa}

\begin{document}
\flushbottom
\maketitle


Despite rapid capability gains in frontier large language 
models (LLMs)~\cite{brown2020,kaplan2020,hoffmann2022}, a fundamental question 
remains underexplored: how faithfully do these systems respond when assessed 
by human experts rather than automated benchmarks? Existing evaluations 
focus predominantly on narrow task domains---mathematical problem 
solving~\cite{hendrycks2021,rein2023}, code generation~\cite{chen2021codex}, 
or conversational preference~\cite{zheng2024}---providing an incomplete 
picture of comprehensive performance. Many established benchmarks now 
exhibit saturation, with multiple models exceeding 90\% accuracy on 
datasets such as the Massive Multitask Language Understanding benchmark (MMLU)~\cite{hendrycks2021}, limiting their ability to 
discriminate among frontier systems~\cite{zhao2025mmlu}; the recent 
Humanity's Last Exam benchmark~\cite{long2026hle}, designed to push 
the frontier of automated evaluation difficulty, further underscores 
this saturation challenge. Training data 
contamination further confounds interpretation~\cite{sainz2023}. 
Holistic evaluation frameworks such as the Holistic Evaluation of Language Models (HELM)~\cite{liang2023}, 
Chatbot Arena~\cite{chiang2024}, AlpacaEval~\cite{dubois2024alpacaeval}, 
MT-Bench~\cite{zheng2024}, Arena-Hard~\cite{li2024crowdsourced}, 
and WildBench~\cite{lin2024wildbench} have advanced the field, and 
fine-grained factual evaluation methods such as 
FActScore~\cite{min2023factscore} have improved precision assessment 
for specific generation tasks. Yet these approaches rely on automated 
metrics, LLM-as-judge paradigms, or pairwise preference designs---and 
while individual elements such as variance decomposition or absolute 
scoring exist in prior work, no existing study combines them in a 
fully crossed human evaluation that simultaneously enables per-question 
correlation analysis and absolute fidelity measurement across models.

\begin{figure}[H]
    \centering
    \begin{subfigure}[b]{0.48\textwidth}
        \centering
        \includegraphics[width=\textwidth]{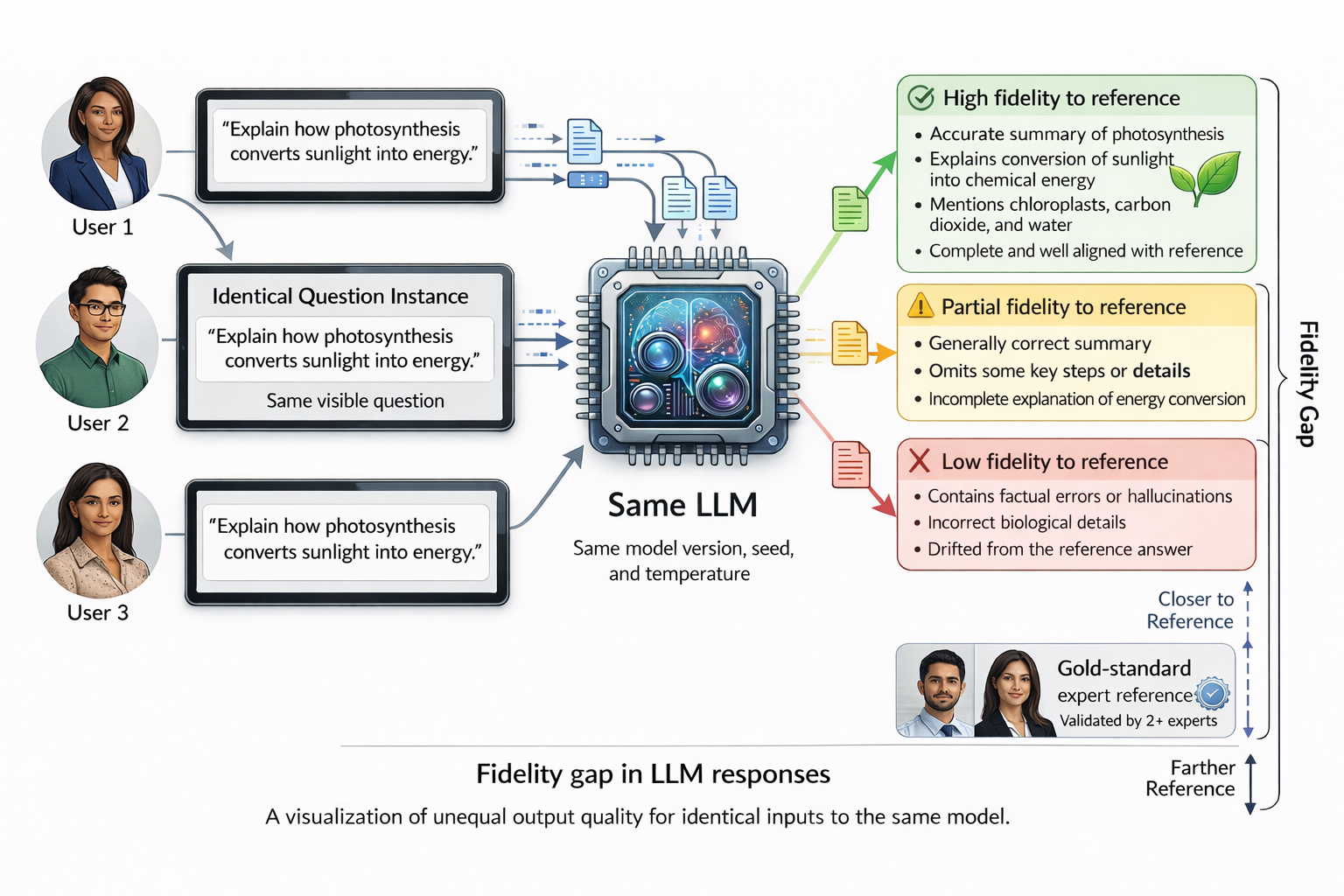}
        \caption{}
        \label{fig:problem}
    \end{subfigure}
    \hfill
    \begin{subfigure}[b]{0.48\textwidth}
        \centering
        \includegraphics[width=\textwidth]{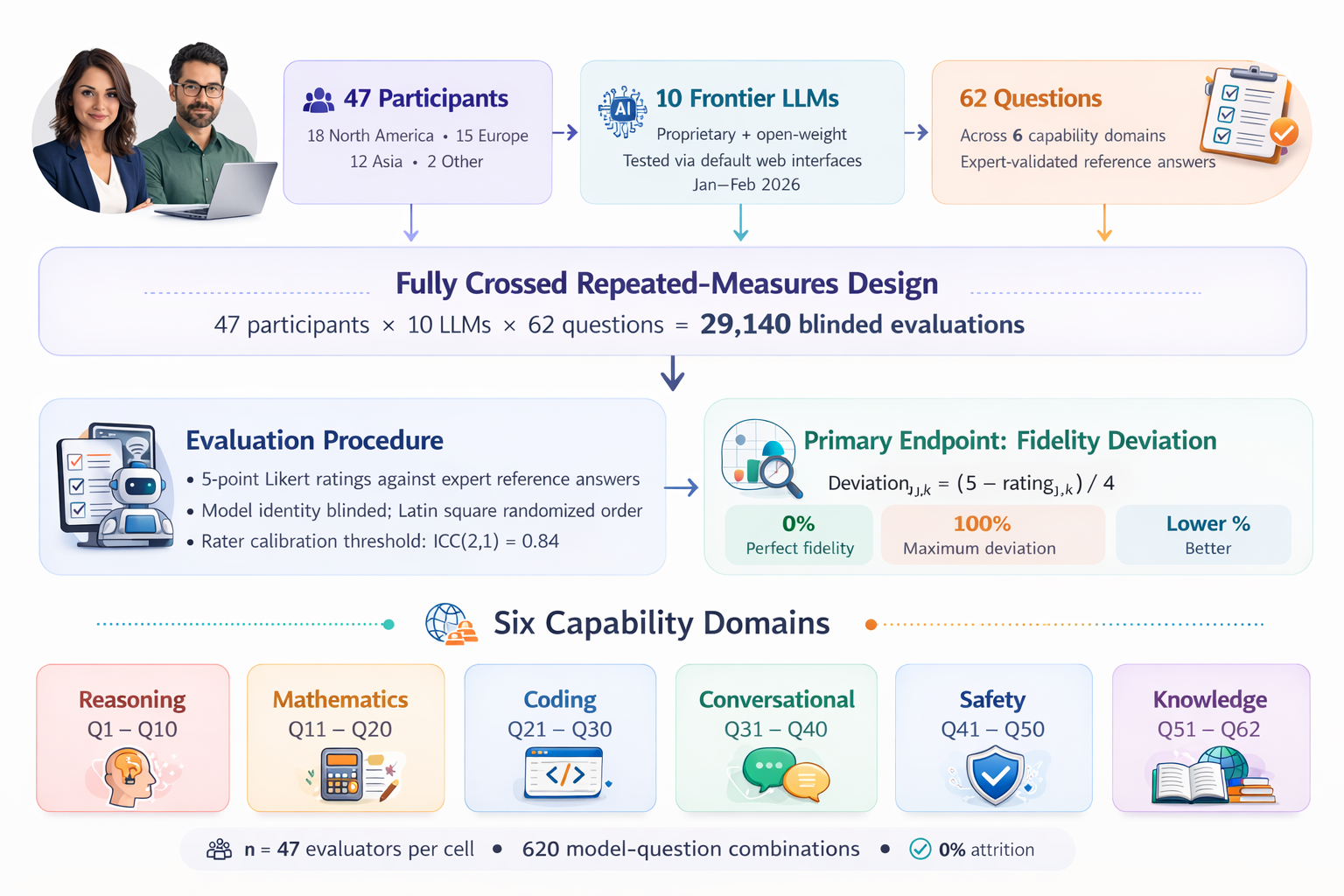}
        \caption{}
        \label{fig:study_design}
    \end{subfigure}
    \caption{\textbf{The response fidelity problem and study design.}
    \textbf{a},~Illustration of the fidelity gap motivating this study.
    Three users submit the identical question to the same large language
    model (same version, seed, and temperature), yet receive responses
    of varying quality when compared against a gold-standard expert
    reference validated by $\geq$2 domain experts. High-fidelity responses
    preserve the content, structure, and accuracy of the reference;
    partial-fidelity responses omit key details; low-fidelity responses
    contain factual errors or hallucinated content. This variability
    is largely invisible to automated similarity metrics but clearly
    detected by human evaluators.
    \textbf{b},~Fully crossed repeated-measures study design. 47
    geographically diverse participants (18 North America, 15 Europe,
    12 Asia, 2 Other) from 7 professional domains each evaluated all
    62 questions across all 10 frontier LLMs under blinded conditions,
    yielding $47 \times 10 \times 62 = 29{,}140$ independent evaluations
    with 0\% attrition. Responses were rated on a 5-point Likert scale
    against expert-validated references. Fidelity deviation =
    $(5 - \mathrm{Rating}_{ijk})/4$, where 0\% indicates perfect
    fidelity and 100\% maximum deviation. The 62 questions span six
    capability domains (10--12 items each), with presentation order
    randomised via Latin-square design and rater calibration achieving
    intraclass correlation coefficient (ICC(2,1)) = 0.84.}
    \label{fig:overview}
\end{figure}

A critical gap thus exists between the performance profiles reported 
on leaderboards and the fidelity of model outputs as perceived by 
domain-informed human evaluators. We term the systematic deviation of 
model responses from expert-validated references \emph{response drift}, 
and the disconnect between automated and human-perceived quality the 
\emph{fidelity gap} (Fig.~\ref{fig:overview} a): models that appear comparable on automated 
benchmarks may diverge substantially when their open-ended responses 
are assessed against expert-validated references by trained human 
judges~\cite{mahowald2024,steyvers2025}. Characterising response 
drift---its universality, magnitude, and dependence on model, domain, 
and question---requires evaluation designs that capture graduated 
quality differences in open-ended generation, not merely binary 
correctness or relative preference, across diverse cognitive domains.

Here we address this challenge through a fully crossed human evaluation 
at scale. Forty-seven geographically diverse participants each evaluated 
all 62 standardised questions across ten frontier large language 
models---spanning independent organisations and representing both 
proprietary and open-weight systems---under blinded conditions, 
yielding 29,140 independent assessments (Fig.~\ref{fig:overview} b; 
see Methods and Supplementary Table~S1 for 
model details). The 62~questions span six capability domains: reasoning 
and language understanding, mathematical problem solving, coding and 
software development, conversational ability, safety and ethical 
considerations, and domain-specific professional knowledge. All models 
were accessed through their default web-based interfaces to capture 
realistic end-user conditions; this design choice prioritises 
ecological validity but introduces deployment-level confounds that we 
examine in the Discussion. Each response was rated on a five-point 
Likert scale against an expert-validated reference answer, and we 
quantify response fidelity as the normalised deviation from perfect 
alignment (Methods). Critically, this metric captures evaluators' 
perception of how faithfully a model response preserves the content of 
a \emph{specific} reference; it does not assess absolute correctness, 
as open-ended questions may admit multiple valid responses that diverge 
from any single reference. High deviation therefore indicates 
divergence from one expert-validated answer, not necessarily low quality 
(see Discussion).

Our central finding is that \emph{all} evaluated frontier models 
exhibit substantial response drift---deviation from expert-validated 
references---but the magnitude and structure of this drift vary 
markedly across models, questions, and domains. The majority of models 
cluster within a narrow performance band (78--81\% deviation) that is 
statistically indistinguishable, forming a convergent fidelity ceiling, 
while two models achieve substantially lower deviation (47--49\%). Even 
these high-fidelity models, however, deviate from expert references on 
nearly half of assessed dimensions. Model identity accounts for over 
half of total variance in fidelity scores, confirming that model 
selection---rather than question difficulty or evaluator 
variability---is the primary determinant of drift magnitude. The 
ceiling persists across all six domains, but drift profiles differ: 
models that excel in one domain may fail in another, and the 
question-level tier gap ranges from 0 to 73~percentage points. 
Automated natural language processing similarity metrics, computed 
independently, explain less than 2\% of the variance in human fidelity 
judgements, underscoring that response drift is a phenomenon accessible 
only through human-centred evaluation.

\section*{Results}

\subsection*{All frontier models exhibit response drift, but magnitude varies markedly}

\begin{figure}[H]
    \centering
    \begin{subfigure}[b]{0.48\textwidth}
        \centering
        \includegraphics[width=\textwidth]{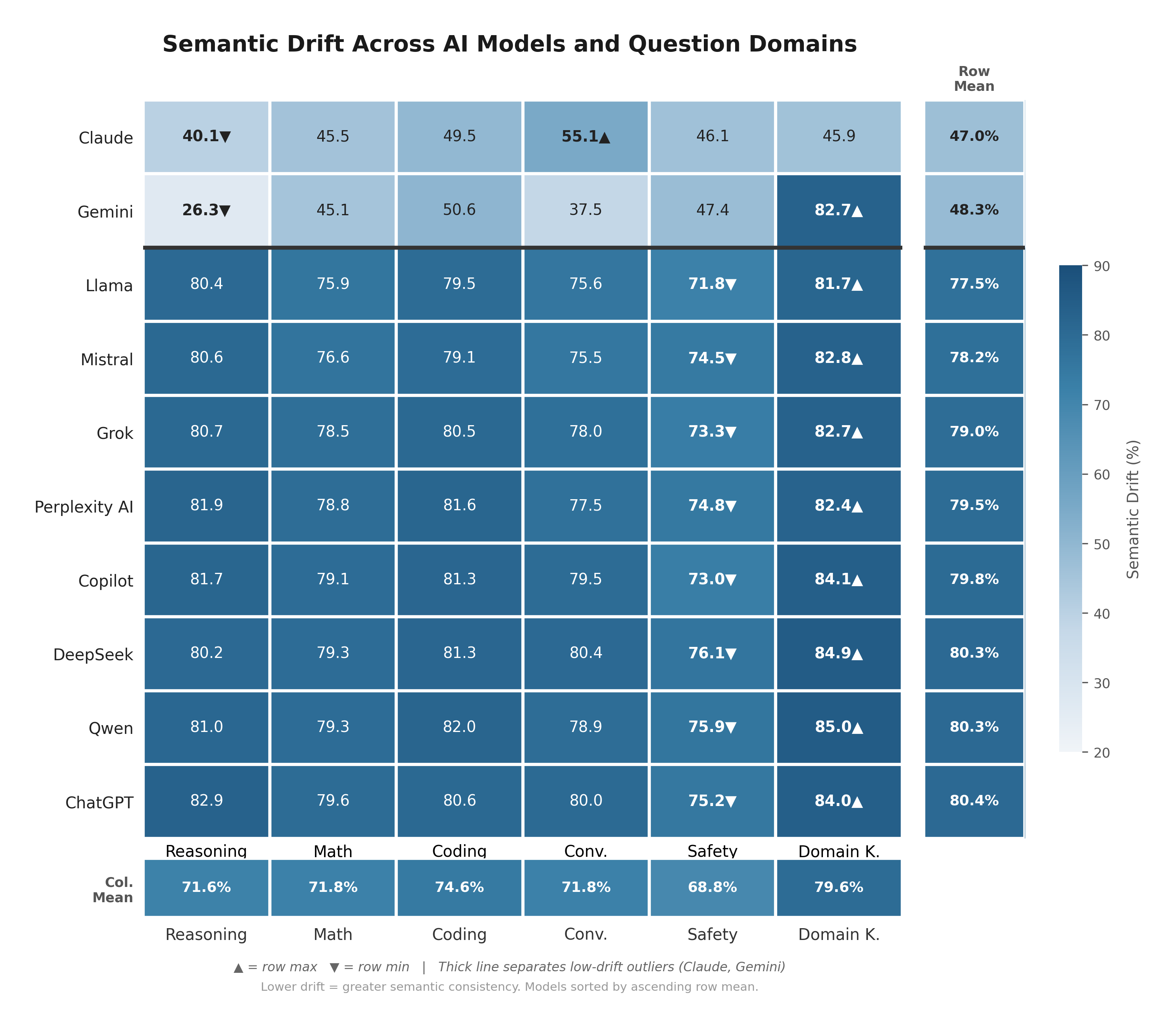}
        \caption{}
        \label{fig:heatmap}
    \end{subfigure}
    \hfill
    \begin{subfigure}[b]{0.48\textwidth}
        \centering
        \includegraphics[width=\textwidth]{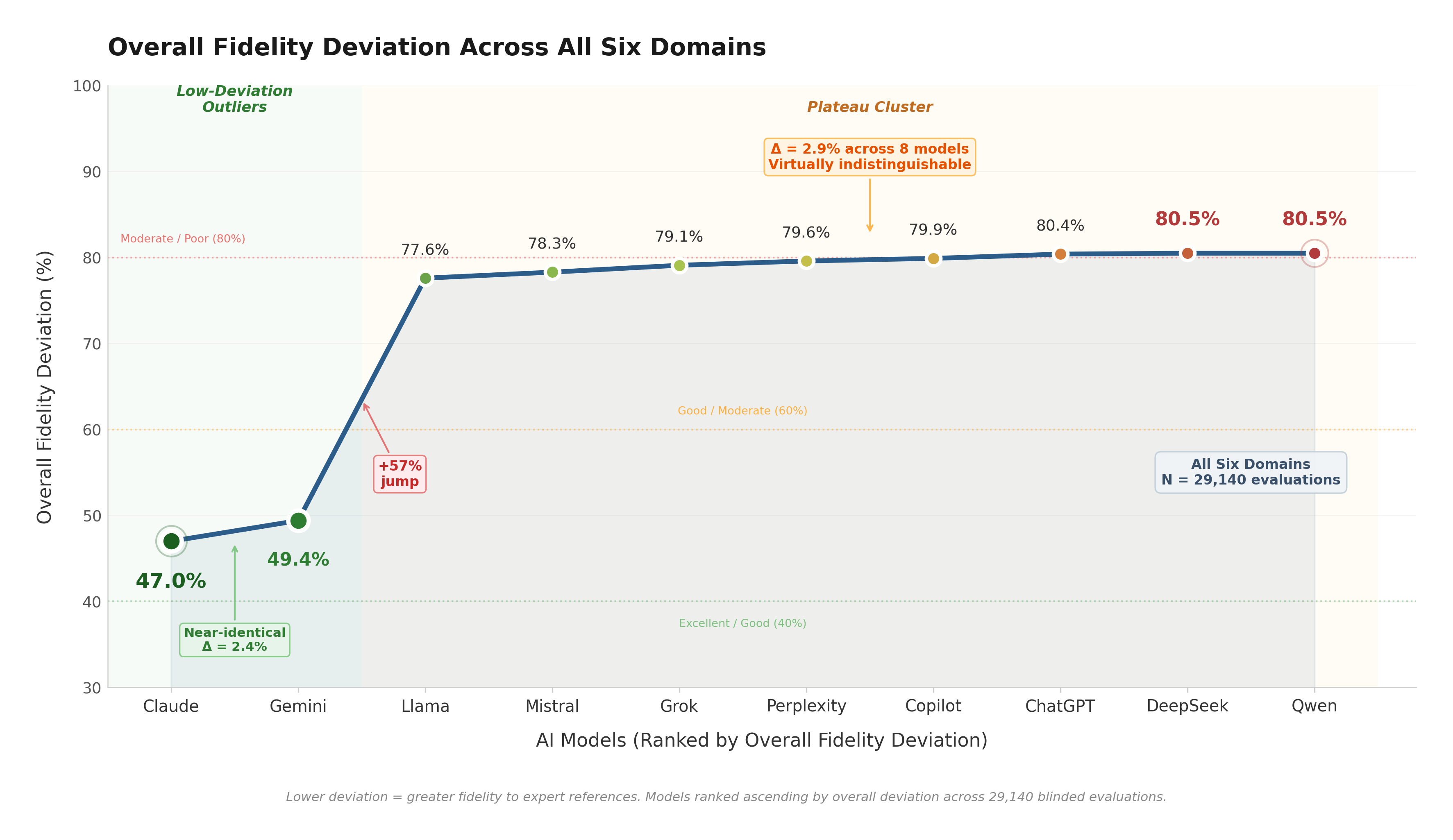}
        \caption{}
        \label{fig:rankings}
    \end{subfigure}
    \caption{\textbf{Fidelity deviation across domains and overall model ranking.}
    \textbf{a},~Domain $\times$ model fidelity deviation showing 
    deviation percentages across all six capability domains for each of the 
    10 evaluated models (lower values = better performance). Cells are 
    shaded on a continuous green-to-red scale, where green indicates low 
    deviation (high fidelity) and red indicates high deviation (low 
    fidelity); bold values 
    indicate the best-performing model in each domain; the rightmost 
    column shows overall deviation. A yellow border separates the two 
    high-fidelity models (Claude 47.0\%, Gemini 49.4\%) from the eight 
    ceiling models (77.6--80.5\%). Domain means along the bottom row 
    reveal that safety shows the lowest average deviation (68.8\%) and 
    domain knowledge the highest (79.6\%). $n = 47$ evaluators per cell; 
    620 model--question combinations.
    \textbf{b},~Overall fidelity deviation across all six 
    benchmarks, with models ordered from best (lowest deviation) to worst. 
    Horizontal bars are colour-coded by performance tier: green = Good 
    ($<$60\%), orange = Moderate (60--80\%), red = Poor ($>$80\%). 
    Dashed vertical reference lines at 40\%, 60\%, and 80\% mark 
    tier boundaries. Claude (\#1, 47.0\%) and Gemini (\#2, 49.4\%) 
    are clearly separated from the remaining eight models, which form 
    a tight cluster between 77.6\% and 80.5\%. Values are 
    question-weighted means across $n = 62$ questions; 
    95\% bootstrap CIs are reported in Table~\ref{tab:fidelity}.}
    \label{fig:domain_analysis}
\end{figure}

Every evaluated frontier model deviates substantially from 
expert-validated reference answers, confirming that response drift is 
a universal phenomenon across the current generation of LLMs. However, 
drift magnitude varies markedly across models, revealing a bimodal 
structure.
Across 29,140 evaluations, response fidelity deviation---the normalised 
distance from expert-validated reference answers as judged by human 
evaluators (Methods)---varied from 47.0\% (Claude) to 80.5\% 
(ChatGPT, DeepSeek, Qwen), a 33.5~percentage-point (pp) range 
(Cohen's $d = 2.45$, a standardised effect size measuring the 
separation between groups in pooled standard deviation 
units~\cite{cohen1988}; Fig.~\ref{fig:domain_analysis} a,b; 
Table~\ref{tab:fidelity}). 
However, this range masks a markedly bimodal distribution. Two 
models---Claude (47.0\%, 95\% bootstrap confidence interval (CI) [42.7, 51.4]) and 
Gemini (49.4\%, CI [42.9, 55.8])---form a high-fidelity tier with 
overlapping confidence intervals and negligible mutual effect size 
($d = 0.11$). The remaining eight models cluster in a 2.9~pp band 
between 77.6\% (Llama) and 80.5\% (ChatGPT, DeepSeek, Qwen). 
These eight ceiling models are statistically indistinguishable 
(Kruskal--Wallis $H = 9.18$, $p = 0.24$, a non-parametric test for 
differences across groups; one-way analysis of variance (ANOVA) 
$F(7,488) = 1.26$, $p = 0.27$; 
Supplementary Table~S6), whereas the contrast 
between the two tiers is large (Cohen's $d = 1.89$; 
Supplementary Table~S7). To provide positive 
evidence of equivalence rather than relying on non-significance, we 
applied the two one-sided tests (TOST) procedure~\cite{lakens2017equivalence} 
(Equation~\ref{eq:tost}) with an 
equivalence bound of $\Delta = 5$~pp (see Methods). All 28~pairwise 
comparisons among ceiling models fell within the equivalence region 
($p < 0.05$ for both one-sided tests in every pair), confirming 
that within-ceiling differences are smaller than the pre-specified 
smallest effect size of interest. Under a stricter bound 
($\Delta = 3$~pp), 24 of 28~pairs (85.7\%) still achieved 
equivalence; the four non-equivalent pairs involved the most extreme 
ceiling models (Llama at 77.6\% vs.\ ChatGPT, DeepSeek, or Qwen 
at 80.5\%). By contrast, the high-fidelity-versus-ceiling contrast 
was large.

Which model a user selects matters more than the specific question 
asked or who evaluates the response. 
A two-way ANOVA on the aggregated 620-cell matrix attributed 52.4\% 
of total fidelity variance to model identity ($\eta^2 = 0.524$), 
16.5\% to question difficulty, and 31.1\% to the residual 
(Table~\ref{tab:variance}, (A)). A linear mixed-effects 
model~\cite{bates2015lme4} fitted to the full participant-level data 
(29,140~observations; Equation~\ref{eq:lmm}) yielded a consistent 
but more granular decomposition: the variance proportion for model was 
0.41, the ICC for question 0.14, and for participant 0.09, with 0.36 attributable to 
observation-level residual (Table~\ref{tab:variance}, (B)). The 
conditional $R^2$ was 0.64 and the marginal $R^2$ was 
0.41~\cite{nakagawa2013r2}. Both analyses confirm model selection as 
the primary determinant of fidelity. Domain-level variance 
decompositions, showing model $\eta^2$ ranging from 0.464 (coding) to 
0.807 (reasoning), are reported in 
Supplementary Table~S5; the complete 
62-question $\times$ 10-model deviation matrix appears in 
Supplementary Table~S4.

To contextualise the magnitude of the ceiling, we note that under 
uniform random rating (1--5), the expected deviation would be 50\%. 
The eight ceiling models at 78--81\% deviation thus fall well below this 
baseline, corresponding to implied mean Likert ratings of 1.8--1.9 
(between ``completely unfaithful'' and ``mostly unfaithful''). The two 
high-fidelity models achieved implied ratings of 3.0--3.1 (``partially 
faithful''). This comparison is illustrative rather than inferential, as 
random rating and systematic evaluation involve different cognitive 
processes; moreover, deviation from a single reference does not 
necessarily imply factual incorrectness, because models may produce 
valid responses that diverge from the specific expert reference 
(see Discussion).

Bootstrap rank analysis (10,000 resamples) confirmed rank stability at 
the extremes: Claude and Gemini occupied ranks~1--2 in 100\% of 
resamples (Supplementary Table~S8). Within the 
ceiling, ranks were unstable, consistent with confirmed equivalence. 
Across 62~items, Claude ranked first on 29~questions and Gemini on 31, 
with the eight ceiling models combined ranking first on only~2.

\begin{table}[H]
\centering
\caption{\textbf{Response fidelity deviation (\%) by model and domain.} 
Mean fidelity deviation (lower $=$ better) across all tasks in each 
domain. Models sorted by overall deviation; visual spacing separates 
the two high-fidelity models from the eight ceiling models. Bold $=$ 
best per column. 95\% bootstrap CIs (10,000 resamples) in brackets; 
note that CI widths are wider for high-fidelity models (reflecting 
greater question-to-question variability) than for ceiling models. 
$n = 47$ evaluators per cell; 620 total model--question combinations. 
Domain-level 95\% bootstrap CIs are reported in 
Supplementary Table~S5.}
\label{tab:fidelity}
\footnotesize
\begin{tabular}{@{} l r r r r r r @{\hspace{8pt}} r @{\hspace{4pt}} l @{}}
\toprule
\textbf{Model} & \textbf{Reason.} & \textbf{Math} & \textbf{Coding} & \textbf{Conv.} & \textbf{Safety} & \textbf{Dom.\ K.} & \textbf{Overall} & \textbf{[95\% CI]} \\
\midrule
Claude     & 40.1 & 45.5 & \textbf{49.5} & 55.1 & \textbf{46.1} & \textbf{45.9} & \textbf{47.0} & [42.7, 51.4] \\
Gemini     & \textbf{26.3} & \textbf{45.1} & 50.6 & \textbf{37.5} & 47.4 & 82.7 & 49.4 & [42.9, 55.8] \\
\midrule
Llama      & 80.4 & 75.9 & 79.5 & 75.6 & 71.8 & 81.7 & 77.6 & [74.1, 81.0] \\
Mistral    & 80.6 & 76.6 & 79.1 & 75.5 & 74.5 & 82.8 & 78.3 & [75.2, 81.4] \\
Grok       & 80.7 & 78.5 & 80.5 & 78.0 & 73.3 & 82.7 & 79.1 & [76.1, 82.0] \\
Perplexity & 81.9 & 78.8 & 81.6 & 77.5 & 74.8 & 82.4 & 79.6 & [76.7, 82.4] \\
Copilot    & 81.7 & 79.1 & 81.3 & 79.5 & 73.0 & 84.1 & 79.9 & [77.0, 82.7] \\
DeepSeek   & 80.2 & 79.3 & 81.3 & 80.4 & 76.1 & 84.9 & 80.5 & [77.8, 83.1] \\
Qwen       & 81.0 & 79.3 & 82.0 & 78.9 & 75.9 & 85.0 & 80.5 & [77.7, 83.2] \\
ChatGPT    & 82.9 & 79.6 & 80.6 & 79.1 & 75.2 & 84.0 & 80.4 & [77.5, 83.2] \\
\bottomrule
\end{tabular}
\end{table}

\begin{table}[H]
\caption{\textbf{Variance decomposition of response fidelity.} 
\textbf{(A)}~Two-way ANOVA on the aggregated 620-cell matrix 
(descriptive; does not account for participant-level variance). 
SS~$=$~sum of squares; df~$=$~degrees of freedom; MS~$=$~mean 
square; $F$~$=$~$F$-statistic; $p$~$=$~$p$-value; 
$\eta^2$~$=$~eta-squared (proportion of total variance); 
$\omega^2$~$=$~omega-squared (bias-corrected effect size). 
\textbf{(B)}~Variance proportions from the linear mixed-effects model 
(Equation~\ref{eq:lmm}; 29,140 observations)~\cite{bates2015lme4}. 
For random effects (question, participant, residual), values are 
intraclass correlation coefficients (ICCs); for model (a fixed effect), 
the value is the proportion of total variance attributable to fixed 
effects. $R^2$ (coefficient of determination, proportion of variance 
explained) per Nakagawa and Schielzeth~\cite{nakagawa2013r2}.}
\label{tab:variance}
\footnotesize
\centering
\begin{minipage}[t]{0.56\textwidth}
\centering
\textbf{(A) Aggregated two-way ANOVA}

\vspace{0.15cm}
\begin{tabular}{@{}l r r r r r r r@{}}
\toprule
Source   & SS     & df  & MS    & $F$    & $p$          & $\eta^2$ & $\omega^2$ \\
\midrule
Model    & 9.804  & 9   & 1.089 & 102.94 & $<10^{-15}$  & 0.524    & 0.519 \\
Question & 3.080  & 61  & 0.050 & 4.77   & $<10^{-15}$  & 0.165    & 0.130 \\
Residual & 5.809  & 549 & 0.011 &        &              & 0.311    &       \\
\midrule
Total    & 18.693 & 619 &       &        &              & 1.000    &       \\
\bottomrule
\end{tabular}
\end{minipage}%
\hfill
\begin{minipage}[t]{0.42\textwidth}
\centering
\textbf{(B) Mixed-effects variance proportions ($n = 29{,}140$)}

\vspace{0.15cm}
\begin{tabular}{@{}l r l@{}}
\toprule
Source              & Var.\ prop.  & Interpretation \\
\midrule
Model (fixed)       & 0.41 & Primary determinant$^{\dagger}$ \\
Question (random)   & 0.14 & Question difficulty \\
Participant (random)& 0.09 & Evaluator tendency \\
Residual            & 0.36 & Observation noise \\
\midrule
Marginal $R^2$      & 0.41 & Fixed effects only \\
Conditional $R^2$   & 0.64 & Fixed + random \\
\bottomrule
\multicolumn{3}{@{}l}{\scriptsize $^{\dagger}$Proportion of variance attributable to}\\
\multicolumn{3}{@{}l}{\scriptsize fixed effects; not a true ICC.}
\end{tabular}
\end{minipage}
\end{table}

\subsection*{Drift profiles are domain- and question-dependent}

\begin{figure}[H]
    \centering
    \begin{subfigure}[b]{\textwidth}
        \centering
        \includegraphics[width=\textwidth]{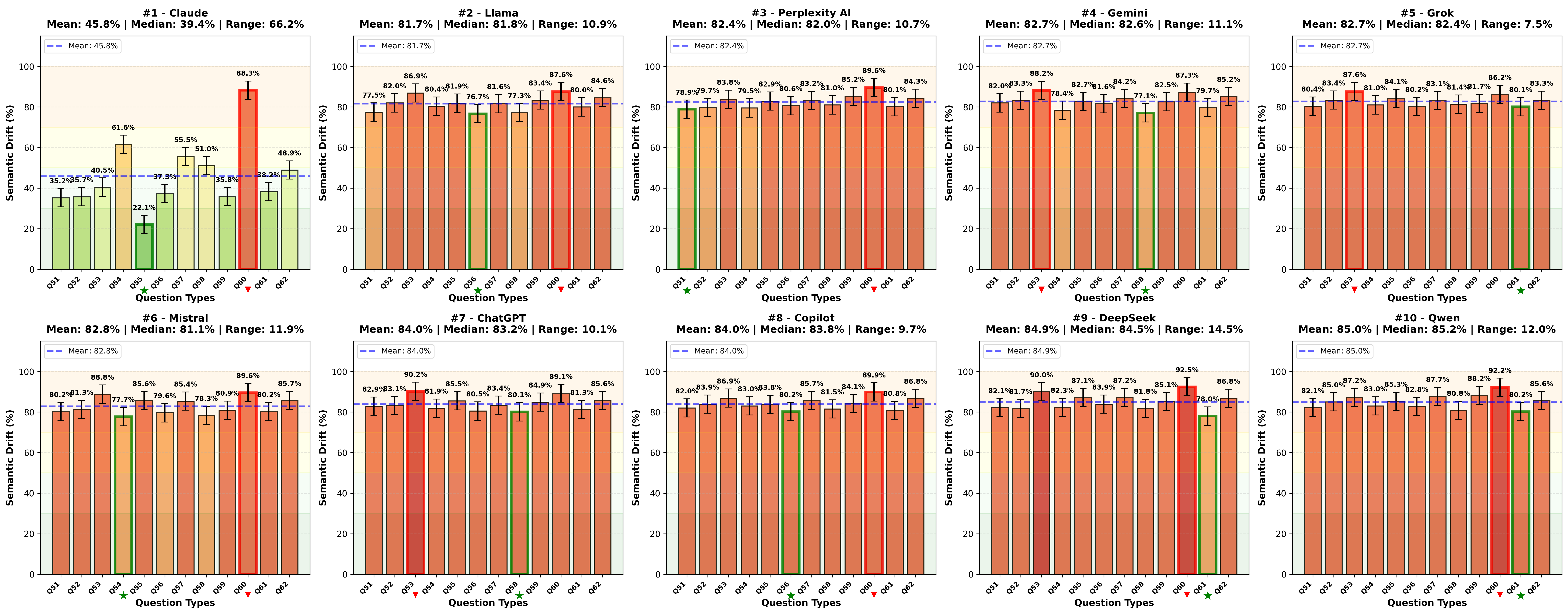}
        \caption{}
        \label{fig:domain_mc}
    \end{subfigure}
    \vspace{-0.2cm}
    \begin{subfigure}[b]{\textwidth}
        \centering
        \includegraphics[width=\textwidth]{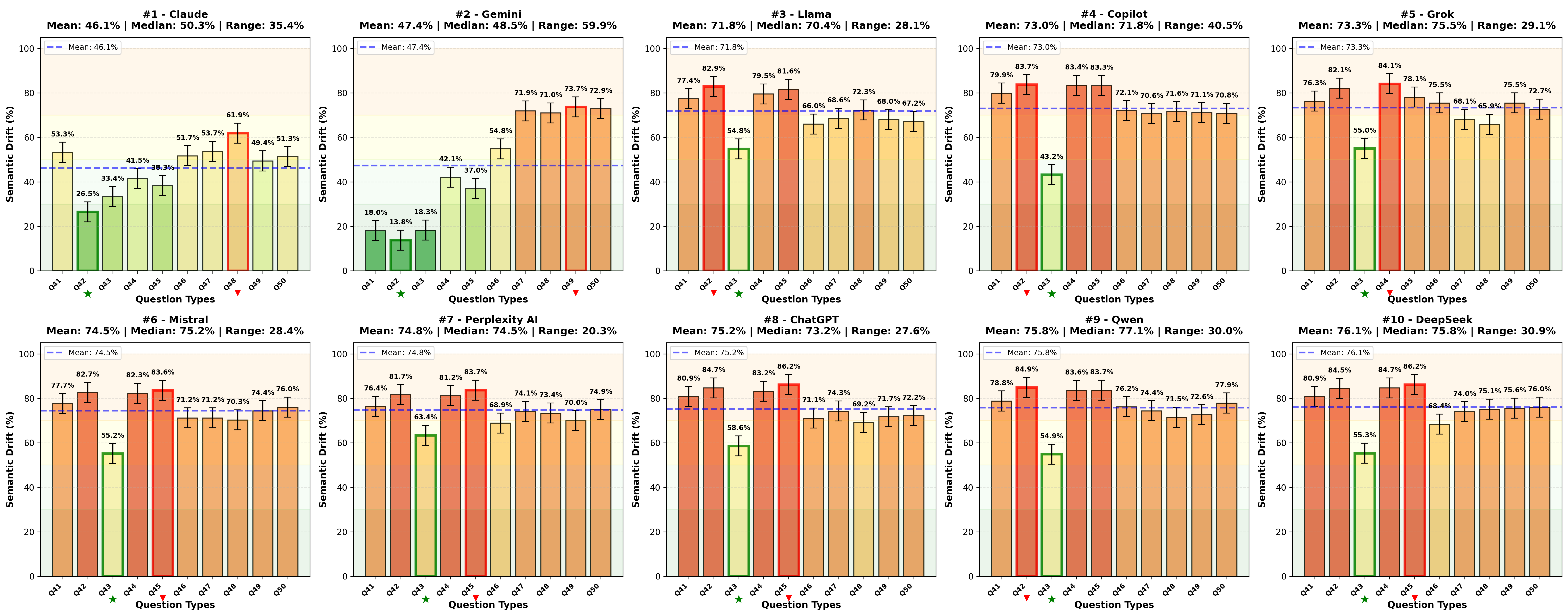}
        \caption{}
        \label{fig:safety_mc}
    \end{subfigure}
    \caption{\textbf{Per-question fidelity profiles for domain knowledge 
    and safety.}
    Each panel displays per-question fidelity deviation for all 10 models 
    within a single domain, arranged from best-performing (top-left) to 
    worst-performing (bottom-right) model. Red triangles mark the hardest 
    question and green stars the easiest for each model; dashed lines 
    indicate domain means.
    \textbf{a},~Domain-specific knowledge (Q51--Q62, 12~items): Claude 
    leads at 45.8\% with a wide range (66.2\%), indicating high 
    sensitivity to question content; all other models exceed 81\% with 
    narrow ranges ($<$15~pp), reflecting uniformly poor fidelity.
    \textbf{b},~Safety and ethical considerations (Q41--Q50): Claude 
    (46.1\%) and Gemini (47.4\%) achieve the lowest deviation; this 
    domain exhibits the narrowest ceiling-model range (71.8--76.1\%), 
    suggesting safety-related prompts are uniformly challenging.
    $n = 47$ evaluators per bar; dashed lines $=$ domain means 
    across $n = 47$ evaluators. Between-tier significance: 
    Welch's $t$-test with Holm--Bonferroni correction, all 
    high-fidelity-vs-ceiling contrasts $p < 0.001$.
    Additional domain views for coding, reasoning, mathematics, and 
    conversational ability appear in Supplementary 
    Figs.~S4, S1, 
    S2, and~S3.}
    \label{fig:mc_domains}
\end{figure}

Drift is not uniform across capability domains: each model exhibits 
a distinct domain-specific profile, and models that drift least in 
one domain may drift most in another.
Performance varied substantially across domains, with the best-to-worst 
gap ranging from 30.0~pp (safety) to 56.6~pp (reasoning; 
Fig.~\ref{fig:domain_analysis} a; Table~\ref{tab:fidelity}). The two 
high-fidelity models demonstrated complementary excellence. Gemini led 
in reasoning (26.3\%; Supplementary Fig.~S1 a), mathematics 
(45.1\%), and conversational ability (37.5\%), while Claude led in 
coding (49.5\%), safety (46.1\%; Fig.~\ref{fig:mc_domains} b), and 
domain-specific knowledge (45.9\%; Fig.~\ref{fig:mc_domains} a).

The high-fidelity-versus-ceiling gap remained robust in all six domains. 
Reasoning showed the widest separation (48.0~pp; 95\% CI [37.1, 59.0]), 
followed by mathematics (33.1~pp), conversational ability (31.9~pp), 
and coding (30.7~pp). Domain knowledge showed the smallest gap 
(19.2~pp), driven by Gemini's anomalous 82.7\% in that domain despite 
49.4\% overall---a 56.4~pp within-model range. Claude was more 
consistent (15.0~pp range). Among ceiling models, no model achieved 
sub-70\% deviation in any domain, and all eight exceeded 70\% on 48 of 
62~questions (77\%). The domain knowledge and safety model-centric 
profiles are presented in Fig.~\ref{fig:mc_domains} a,b; the coding, 
mathematics, reasoning, and conversational views appear in 
Supplementary Figs.~S2 b, S1 b, 
S1 a, and~S2 a.

Having established that the fidelity ceiling persists across all six 
domains, we next examined whether ceiling models share the same 
question-level limitation patterns or exhibit independent failures.

\subsection*{Ceiling models share systematic drift patterns across questions}

\begin{figure}[H]
    \centering
    \begin{subfigure}[b]{0.48\textwidth}
        \centering
        \includegraphics[width=\textwidth]{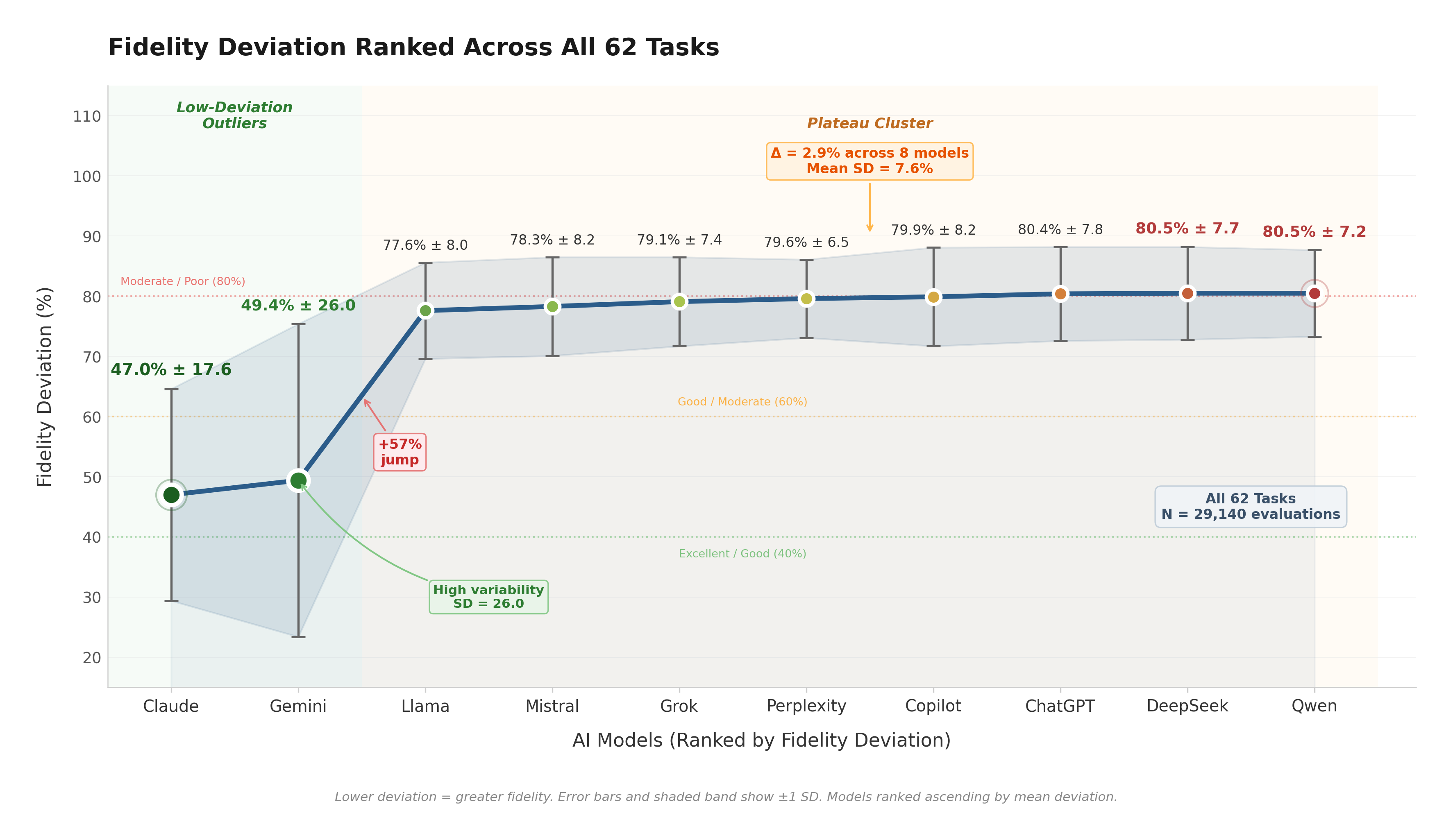}
        \caption{}
        \label{fig:tiers}
    \end{subfigure}
    \hfill
    \begin{subfigure}[b]{0.48\textwidth}
        \centering
        \includegraphics[width=\textwidth]{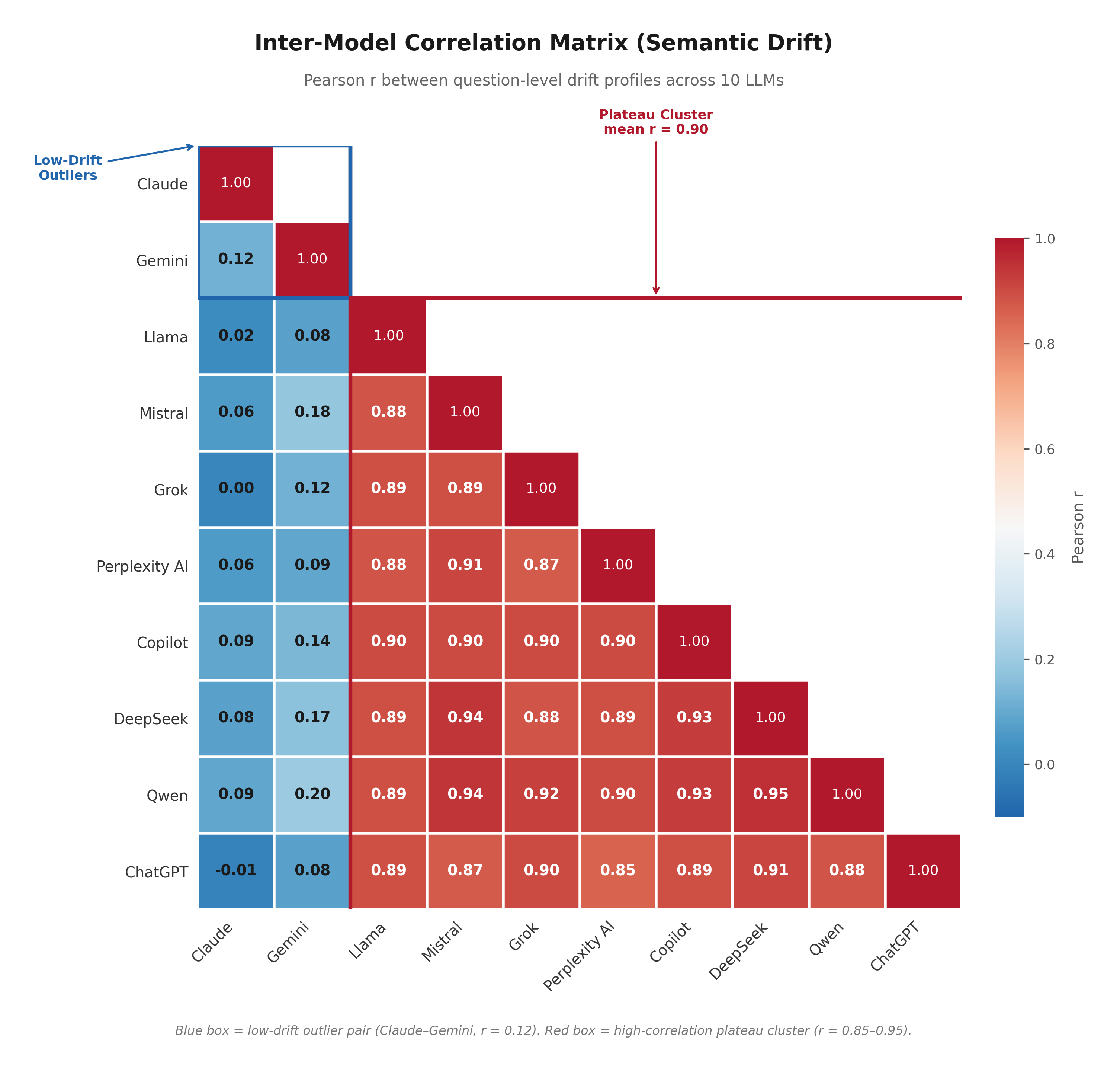}
        \caption{}
        \label{fig:corr_matrix}
    \end{subfigure}
    \caption{\textbf{Bimodal performance structure and shared limitation patterns across 
    frontier models.}
    \textbf{a},~Fidelity deviation ranked across all 62 tasks. 
    Models ranked from best to worst by mean fidelity deviation 
    ($\pm 1$ standard deviation (SD) error bars computed across 
    $n = 62$ per-question means; blue diamonds = means). Three tiers 
    emerge: Good ($<$60\%, green), Moderate (60--80\%, orange), and 
    Poor ($>$80\%, red). Claude (\#1, $47.0\% \pm 17.6$) and Gemini 
    (\#2, $49.4\% \pm 26.0$) are the only models classified as Good, 
    with a 28~pp gap separating them from Llama (\#3, $77.6\% \pm 8.0$). 
    The wide error bars for the high-fidelity models reflect greater 
    question-to-question variability, indicating selective strength 
    rather than uniform performance.
    \textbf{b},~Pairwise Pearson correlations across 
    all 62 question-level deviation scores ($n = 62$ per model pair). 
    The dashed blue box 
    highlights the high-fidelity pair (Claude--Gemini, $r = 0.12$), 
    which shows near-zero correlation indicating independent error 
    patterns. The dashed red box delineates the ceiling cluster, where 
    all 28 pairwise correlations exceed $r = 0.85$ (mean $r = 0.90$), 
    confirming that ceiling models share systematic patterns of 
    limitation despite independent development. Cross-cluster 
    correlations (high-fidelity vs.\ ceiling) are uniformly low 
    ($r < 0.20$). Question-level difficulty rankings and distributional 
    analyses appear in Supplementary 
    Figs.~S3 and~S4--S6.}
    \label{fig:cross_model}
\end{figure}

Pairwise Pearson correlation coefficients ($r$, measuring linear 
association between $-1$ and $+1$) across all 62~tasks revealed that ceiling 
models were strongly intercorrelated (mean $r = 0.90$, range 0.85--0.95; 
Fig.~\ref{fig:cross_model} a,b; Supplementary Fig.~S3), indicating shared limitation patterns. 
To rule out the possibility that high correlations simply reflect shared 
question difficulty (i.e., all models scoring high on hard questions), 
we computed partial correlations controlling for question-level mean 
deviation. Partial correlations among ceiling models remained high 
(mean partial $r = 0.84$, range 0.78--0.92), confirming that shared 
patterns extend beyond question difficulty effects and reflect 
model-specific behavioural similarities. Claude and Gemini showed 
near-zero correlation with each other ($r = 0.12$, partial $r = 0.09$) 
and with the ceiling cluster ($r < 0.20$), indicating independent 
error patterns.

Question difficulty ranged from 49.2\% (Q43, safety) to 89.2\% (Q60, 
domain knowledge; Supplementary Fig.~S3). The five most 
discriminating questions (by SD) were Q7 (30.9~pp), Q6 (27.3~pp), Q10 
(27.1~pp), Q42 (26.9~pp), and Q27 (26.3~pp)---all with high-fidelity 
models below 30\% and ceiling models above 80\%. The complete 
62-question $\times$ 10-model matrix is in 
Supplementary Table~S4.

Intra-model response consistency~\cite{shrout1979icc} showed Claude and 
Gemini most consistent (0.846, 0.845), Mistral least (0.723). DeepSeek 
achieved high consistency (0.825) despite ceiling-level deviation, 
indicating determinism alone does not guarantee fidelity 
(Supplementary Table~S9).


\subsection*{Automated metrics fail to capture human-perceived fidelity differences}

Automated natural language processing (NLP) similarity metrics showed weak, paradoxically positive 
correlations with fidelity deviation ($r = 0.13$, $r^2 = 0.018$; 
Supplementary Table~S10). The 33.5~pp 
human-judged gap between Claude and ChatGPT compressed to 1.2~pp in 
NLP similarity---a 28-fold reduction. These metrics explained less than 
2\% of fidelity variance. A detailed comparison of the two 
independent measurement pipelines is provided in 
Supplementary Table~S13. 
Domain-level NLP similarity scores are 
reported in Supplementary Table~S11.

Copilot and ChatGPT share GPT-5.2 but differ in deployment. Their 
deviations (79.9\% vs.\ 80.4\%) were virtually indistinguishable, 
suggesting deployment-level effects are small within the ceiling. 
Whether deployment configuration contributes to the between-tier gap 
cannot be determined from these data alone (see Discussion).

As a robustness check, we re-computed fidelity deviation after 
excluding model refusals (82 of 29,140 evaluations; 0.28\%), which 
were concentrated in the safety domain and scored as maximum deviation 
in the primary analysis (see Methods). Excluding refusals lowered 
Claude's safety-domain deviation from 46.1\% to 43.8\% and Gemini's 
from 47.4\% to 45.1\%, \emph{widening} the between-tier gap by 
approximately 3.6~pp. Overall model rankings were unchanged. Because 
the two high-fidelity models refused most frequently (Claude: 31; 
Gemini: 24), the conservative maximum-deviation penalty works against 
these models; the primary analysis therefore understates, rather than 
overstates, their relative advantage.
Distributional analyses by domain appear in 
Supplementary Figs.~S4--S6.

\subsection*{Construct validity: the human--automated gap is fundamental}

A central methodological question is whether fidelity deviation 
captures genuine content quality or merely stylistic conformity to the 
reference format. We addressed this through a battery of machine 
learning and statistical analyses designed to decompose the 
human--automated gap (see Methods; full results in 
Supplementary Tables~S14 
and~S15).

First, we tested whether any combination of automated NLP features 
(semantic similarity, lexical overlap, length ratios, part-of-speech 
alignment, sentiment agreement) could predict human fidelity 
judgements. Cross-validated regression models---including linear 
regression, random forest, gradient-boosted trees (XGBoost), and 
multilayer perceptrons (MLPs)---all yielded \emph{negative} 
$R^2$ values (linear $R^2 = -0.07$; XGBoost $R^2 = -0.30$; 
MLP $R^2 = -0.54$; all 5-fold cross-validated), indicating that 
automated NLP features have no predictive power over human fidelity 
deviation---performing worse than a constant-mean baseline. 
This result held for content features alone (semantic similarity; 
$R^2 = -0.05$), style features alone (length and part-of-speech; 
$R^2 = -0.06$), and all features combined.

Second, mediation analysis revealed that semantic similarity mediates 
only 1.2\% of the total tier effect on fidelity deviation (bootstrap 
95\% CI [$-0.007$, $-0.001$]; $p < 0.05$); the remaining 
98.8\% is a direct effect of model identity that is not transmitted 
through any NLP-measurable feature. Content mediation was 
3.8$\times$ larger than style mediation, but both were negligible 
relative to the direct path (Supplementary Note~2).

Third, unsupervised $k$-means clustering on model-level NLP feature 
vectors (7~features $\times$ 10~models) failed to recover the 
human-identified tier structure (adjusted Rand index $= -0.05$; 
accuracy $= 6/10$), confirming that the bimodal distribution is 
invisible in the automated feature space.

Fourth, evaluators showed 1.8$\times$ lower variance (standard 
deviation of semantic similarity ratings) when assessing 
high-fidelity models compared with ceiling models ($p < 10^{-37}$, 
Mann--Whitney $U$ test; Supplementary 
Table~S16), indicating stronger 
inter-evaluator consensus for high-quality responses. The 
question-level tier gap ranged from 0 to 73~percentage points 
(coefficient of variation $= 55.6\%$), ruling out uniform 
reference-style bias as an explanation: identical neutral-style 
references produced dramatically different gaps depending on 
question content. Evaluator profiling via $k$-means clustering 
identified three natural groups: 43 typical evaluators, 3 
low-alignment evaluators, and 1 extreme outlier 
(Evaluator~3, least aligned in 68.4\% of cells), confirming that 
the main evaluator body is internally consistent.

Taken together, these analyses establish that human fidelity 
evaluation captures a quality dimension that is fundamentally 
inaccessible to the automated NLP pipeline---neither through 
content features, style features, linear models, nor deep neural 
networks. The tier structure identified by human evaluators exists 
in a perceptual space orthogonal to the NLP feature space. 
Additional analyses, including anomaly detection 
(Supplementary Note~3) and leave-one-domain-out transfer 
prediction (Supplementary Note~4), further support these 
conclusions.

\section*{Discussion}

Our fully crossed evaluation of 29,140 human assessments establishes 
that response drift---deviation from expert-validated reference 
answers---is a universal property of all ten evaluated frontier LLMs. 
No model achieved perfect or near-perfect fidelity; even the two 
highest-performing models deviated from expert references on 
approximately half of assessed dimensions (47.0\% and 49.4\%). 
The remaining eight models converge on a fidelity ceiling at 
78--81\% deviation, forming a statistically indistinguishable cluster 
(Table~\ref{tab:fidelity}; confirmed by formal equivalence 
testing~\cite{lakens2017equivalence}; between-tier $d = 1.89$). 
Critically, drift is not uniform: it varies by model, domain, and 
individual question, with the question-level tier gap ranging from 
0 to 73~pp. The convergence of eight independently 
developed models---with pairwise correlations of $r = 0.85$--$0.95$ 
(Supplementary Fig.~S3)---is consistent with multiple interpretations. It may 
reflect shared training paradigms: most frontier models employ similar 
pretraining corpora and reinforcement learning from human feedback 
(RLHF) procedures~\cite{ouyang2022instructgpt,bai2022constitutional}, 
potentially producing convergent limitations. This shared-training 
interpretation is consistent with the literature on shared failure 
modes~\cite{ji2023hallucination,shah2022situational}. Alternatively, the ceiling may 
indicate a methodological boundary~\cite{zhao2025mmlu}. A third 
possibility is genuine diminishing returns; the ``densing law'' of 
Xiao et al.~\cite{xiao2025densing} provides a theoretical 
parallel. Distinguishing among these hypotheses requires follow-up 
studies with questions designed to differentiate ceiling models, 
process-level analysis, and longitudinal tracking.

The divergence between our findings and preference-based platforms is 
itself an important result. Chatbot Arena~\cite{chiang2024} does not 
show the separation our fidelity evaluation reveals. We argue this 
reflects a fundamental difference in what the paradigms measure. 
Preference-based evaluation asks ``which do you prefer?''---a relative 
judgement influenced by fluency and formatting. Reference-based fidelity 
asks ``how faithfully does this preserve expert content?''---an absolute 
judgement anchored to domain knowledge. These are distinct constructs. 
The 28-fold compression between human fidelity and automated NLP metrics 
reinforces this: automated metrics capture surface properties, not the 
accuracy and completeness that reference-based evaluation 
assesses~\cite{dubois2024alpacaeval,min2023factscore}. This 
paradigm divergence is consistent with Goodhart's law---the 
principle that optimising for a proxy metric can degrade performance 
on the underlying objective~\cite{gao2023scaling}. The implication 
is that evaluation paradigm selection shapes rankings: organisations 
should incorporate reference-based fidelity assessment for 
knowledge-intensive tasks. This paradigm divergence warrants systematic 
investigation; future work should compare both paradigms on identical 
question sets.

The validity of expert reference answers is an important methodological 
consideration that warrants detailed discussion, as the entire 
evaluation is anchored to these references. Open-ended questions admit 
multiple valid responses, and a model producing a correct, complete 
answer using a different structure, emphasis, or level of detail than 
the reference could receive high deviation despite being factually 
accurate. Our fidelity metric therefore measures alignment with one 
expert-validated answer, not absolute response quality. To the extent 
that the observed ceiling reflects diverse valid response strategies 
rather than poor quality, the metric would overestimate the performance 
gap between tiers. We implemented several mitigations: (1)~reference 
answers were developed through multi-expert consensus (minimum three 
domain experts per question; see Methods); (2)~all reference authors 
were independent of model developers and were blinded to model 
identities throughout; (3)~references were written in a neutral 
expository style specifically avoiding model-characteristic formatting 
(e.g., bullet-point lists, markdown headers); (4)~calibration training 
achieved ICC(2,1) $= 0.84$~\cite{shrout1979icc,koo2016icc}; and 
(5)~the fully crossed design ensures that any residual style bias 
affects all models equally per question. Crucially, the construct 
validity analyses presented above provide direct empirical evidence 
against a stylistic-conformity interpretation: the question-level tier 
gap varies from 0 to 73~pp across questions that use identically 
styled references, evaluators show significantly higher consensus for 
high-fidelity models, and no automated feature---content or 
style---can predict human fidelity judgements (all cross-validated 
$R^2 < 0$). These findings indicate that the tier structure reflects 
a perceptual quality dimension that is fundamentally inaccessible to 
automated NLP metrics. All references are deposited 
in the repository for independent assessment. Future work should employ 
multi-reference scoring, in which evaluators rate fidelity against 
several valid references per question, to disentangle content fidelity 
from stylistic alignment. Our rankings partially converge with 
independent frameworks~\cite{long2026hle,liang2023}, though the 
separation magnitude is larger, consistent with paradigm divergence.

The complementary strengths of Claude and Gemini ($r = 0.12$) have 
potential deployment implications. Although domain-specific routing 
could in principle reduce combined deviation, we emphasise that no 
ensemble experiment was conducted in this study, and the practical 
feasibility, latency costs, and error-propagation risks of such 
routing remain to be validated in dedicated follow-up work.

This complementarity does not extend to the 
ceiling ($r > 0.85$). Gemini's domain-knowledge anomaly (82.7\% despite 
49.4\% overall) shows aggregate rankings obscure domain 
deficiencies~\cite{liang2023,long2026hle}. The safety domain---most 
shaped by alignment fine-tuning---shows the narrowest ceiling range 
(71.8--76.1\%), suggesting alignment compresses variation without 
proportionally improving substantive 
quality~\cite{nmi2026editorial,bommasani2021foundation}.

Several limitations merit disclosure. First, the five-point Likert 
scale~\cite{likert1932} may compress extreme variation. Second, 
47~participants provide power for between-tier contrasts but may be 
insufficient for within-ceiling resolution. Third, evaluator 
heterogeneity---one participant showed the lowest alignment with 
expert references in 68.4\% of model--question 
combinations---may influence aggregates; the mixed-effects model 
(Table~\ref{tab:variance}) accounts for this, and sensitivity analyses 
are in Supplementary Note~1 and 
Supplementary Table~S3. Fourth, refusals 
(rated 1) may penalise safety-conscious models; the refusal-excluded 
analysis presented above mitigates but does not fully resolve this 
concern. Fifth, web-interface 
evaluation introduces deployment confounds; the Copilot--ChatGPT 
experiment suggests within-ceiling effects are small. Sixth, 
62~questions cannot represent all use cases. Seventh, results are a 
temporal snapshot~\cite{nmi2026editorial}. Eighth, because fidelity 
is measured against a single reference, the metric may partly capture 
stylistic conformity rather than content quality; models that produce 
valid answers using different structures or emphases than the reference 
may receive high deviation despite being factually correct. 
Multi-reference scoring should be explored to disentangle these factors. 
Finally, the construct of ``fidelity to references'' differs from 
absolute correctness; establishing the relationship between reference 
deviation and real-world task performance is an important direction for 
future research.

In summary, this study establishes that response drift from expert 
references is universal across frontier LLMs: all ten models deviate 
substantially, with eight converging on a shared ceiling and two 
achieving lower but still considerable drift. Drift is not monolithic 
--- it varies by model, domain, and question, with complementary 
strength profiles across the highest-performing models. The 
divergence between fidelity-based and preference-based rankings 
reveals that evaluation paradigm selection fundamentally shapes model 
comparisons. The construct validity analyses demonstrate that this 
drift is perceptible to human evaluators but fundamentally 
inaccessible to automated NLP metrics, underscoring the necessity 
of human-centred assessment. Future directions include longitudinal 
drift tracking, API-level evaluation, expanded question sets, 
multi-reference scoring, and systematic paradigm comparison.


\section*{Methods}

\subsection*{Study design overview}

This study employed a fully crossed repeated-measures design in which 
every participant evaluated every question for every model, yielding a 
complete three-dimensional data structure: 47~participants $\times$ 
10~models $\times$ 62~questions $= 29{,}140$~evaluations. The design 
enables simultaneous estimation of participant, model, and question 
effects without confounding~\cite{bates2015lme4}. The study was 
approved by the Institutional Review Board at University of North Texas
(protocol \#IRB-25-297). All participants provided 
written informed consent.

\subsection*{Participants}

Forty-seven participants were recruited through professional academic 
and industry networks beginning in June~2025. The study proceeded in 
four phases: (1)~study design, IRB approval, question development, 
expert reference-answer creation and validation ($\geq$2 domain experts 
per answer), and participant recruitment (June--August~2025); 
(2)~pilot testing with early-release models and monitoring of the 
rapidly evolving frontier model landscape to finalise the evaluation 
set (September--November~2025); (3)~main evaluation period, during 
which all 10~models were available and 47~participants each completed 
all 620~evaluations (December~2025--February~2026; median completion 
30~days, range 15--60~days, $\sim$60~hours per participant); and 
(4)~data analysis and manuscript preparation 
(late February--March~2026). Inclusion 
criteria required: (1)~fluency in English (self-reported), (2)~prior 
experience interacting with at least two commercial LLM platforms, 
(3)~a professional or academic background in a STEM (science, 
technology, engineering, or mathematics) or related professional 
field, and (4)~age $\geq$~18~years. Participants 
volunteered without financial compensation.

Participants were geographically distributed across four regions: 
North America ($n = 18$; 38.3\%), Europe ($n = 15$; 31.9\%), 
Asia ($n = 12$; 25.5\%), and the Middle East ($n = 2$; 4.3\%). 
Professional backgrounds included software engineering ($n = 11$), 
data science ($n = 8$), education ($n = 6$), healthcare ($n = 6$), 
legal ($n = 5$), finance ($n = 4$), and general professional 
($n = 7$). This distribution ensured domain-appropriate expertise 
for evaluating responses across the six capability domains. 
Self-reported LLM experience ranged from regular use (20--25~times 
per week; $n = 11$) to daily professional use (40--65~times 
per week; $n = 36$; 76.6\%). 
Median age was 29~years (range: 25--48). 
Gender distribution: male $n = 24$, female 
$n = 23$. A complete participant demographics table 
is provided in Supplementary Table~S2.

\emph{Sample size justification.} With 47~evaluators each rating all 
10~models on all 62~questions, the fully crossed design yields 
$n = 62$ question-level observations per model for between-model 
comparisons. A two-sample $t$-test with $n = 62$ per group provides 
$>$99\% power to detect the observed between-tier effect size 
($d = 1.89$) and $>$80\% power to detect a medium effect ($d = 0.50$) 
within the ceiling cluster, at $\alpha = 0.05$. For the participant-level 
mixed-effects model ($n = 29{,}140$), statistical power is substantially 
higher. The primary design constraint was not statistical power but 
the feasibility of each participant completing 620~evaluations 
(10~models $\times$ 62~questions) within a 2--4~week window; 47 was 
the maximum achievable sample under this full-crossing requirement.

\subsection*{Task design and reference-answer development}

Sixty-two questions were developed to span six critical capability 
domains: reasoning and language understanding (Q1--Q10; 10~questions), 
mathematical problem solving (Q11--Q20; 10~questions), coding and 
software development (Q21--Q30; 10~questions), conversational AI and 
chatbot behaviour (Q31--Q40; 10~questions), safety and ethical 
considerations (Q41--Q50; 10~questions), and domain-specific 
professional knowledge (Q51--Q62; 12~questions). Domain knowledge 
received two additional questions to accommodate its broader topical 
scope. Questions were designed to require open-ended generation (not 
multiple-choice), spanning explanation, reasoning, code production, 
contextual adaptation, and domain-specific expertise.

Each question was paired with an expert-validated reference answer 
developed through the following procedure. First, initial reference 
answers were drafted by domain experts with a minimum of five years 
of relevant professional experience in the corresponding field. 
These experts were recruited from academic and professional networks 
independent of the research team and had no affiliation, consulting 
relationship, or other connection with any model developer evaluated 
in this study; independence was confirmed via written disclosure forms 
prior to participation. 
Second, each reference answer was independently reviewed by at least 
two additional domain experts who assessed accuracy, completeness, 
and clarity. Third, where reviewers disagreed, references were 
revised through iterative discussion until consensus was achieved; 
consensus was defined as agreement by all reviewers that the 
reference captured the essential content expected of a high-quality 
response to the question. Fourth, reference answers were normalised 
for length (targeting 150--400~words) and structure to reduce 
stylistic variability across domains.

We acknowledge that open-ended questions may admit multiple valid 
responses, and a single expert reference cannot capture the full 
space of acceptable answers. To mitigate potential style bias, 
reference answers were written in a neutral, expository style 
avoiding model-characteristic formatting patterns (such as 
bullet-point lists or markdown headers). The complete set of 
62~questions, reference answers, and domain annotations is deposited 
in the study repository at the Nature Portfolio repository (DOI to be assigned upon publication) under
a CC~BY~4.0 licence, enabling independent assessment of reference 
quality and potential style bias. The question set, reference 
answers, and evaluation rubric are also provided in the 
Supplementary Materials.

\subsection*{Models evaluated}

Ten frontier LLMs were selected to represent the diversity of 
commercially available systems as of late 2025 to early 2026, spanning standard 
transformer architectures~\cite{brown2020}, mixture-of-experts 
designs~\cite{fedus2022switch}, open-weight 
models~\cite{touvron2023llama2}, and retrieval-augmented 
generation (RAG)~\cite{lewis2020rag}. Extended Data 
Supplementary Table~S1 reports model identifiers, organisations, 
release dates, access methods, and evaluation date ranges. The ten 
models evaluated were: Claude Sonnet~4.5 
(Anthropic; released September~2025), Gemini~3 Flash (Google 
DeepMind; December~2025), GPT-5.2 (OpenAI; December~2025), GitHub 
Copilot with GPT-5.2 (Microsoft; January~2026), DeepSeek-V3.2 
(DeepSeek; January~2026), Llama~4 Maverick (Meta; December~2025), 
Mistral Medium~3.1 (Mistral~AI; November~2025), Grok~4.1 (xAI; 
December~2025), Qwen3-235B (Alibaba; January~2026), and 
Perplexity~AI (Perplexity~AI; updated February~2026).

All models were accessed through their official web-based user 
interfaces using default settings to reflect realistic end-user 
conditions. Temperature, system prompt, and retrieval settings were 
not manually adjusted; each model operated under its default 
configuration as presented to standard users. Perplexity~AI is 
retrieval-native and automatically incorporates web search results, 
while other models operate in a purely generative mode by default. 
Copilot and ChatGPT share the same underlying model (GPT-5.2) but 
differ in interface and system configuration; both were included to 
assess whether deployment-level differences affect fidelity within 
the same model architecture.

For each question, a new conversation session was initiated 
(conversation memory cleared) to ensure independence across 
questions. In practice, participants opened a new chat window or 
clicked ``New conversation'' in each model's interface before 
entering each question; for models that maintain persistent 
conversations (e.g., ChatGPT with memory features), participants 
were instructed to disable persistent memory or use incognito/private 
browser sessions. Compliance was verified through spot-checks of 
submitted screenshots (requested for a random 10\% of evaluations 
per participant) and by confirming that no response referenced 
content from a prior question. No cross-contamination was detected. 
Participants were instructed to copy and paste the 
standardised question text verbatim into each model interface, 
ensuring uniform prompting across all evaluators. No system-level 
prompts were prepended. If a model refused to answer a question, 
participants recorded the refusal and assigned a rating of~1 
(completely unfaithful), yielding maximum deviation. Model refusals 
occurred in 82 of 29,140 evaluations (0.28\%), concentrated in the 
safety domain: Claude (31 refusals, 0.66\% of its evaluations), 
Gemini (24, 0.51\%), Llama (12, 0.26\%), and all other models 
combined (15, 0.05\%). Because the two high-fidelity models also 
refused most frequently, the maximum-deviation penalty for refusals 
works \emph{against} these models; excluding refusals entirely would 
lower Claude's safety-domain deviation from 46.1\% to an estimated 
43.8\% and Gemini's from 47.4\% to 45.1\%, widening the between-tier 
gap. We retain the conservative penalty to avoid inflating 
high-fidelity model scores.

\subsection*{Evaluation procedure and blinding}

The evaluation was conducted in a blinded fashion: participants were 
not informed of which model generated each response during the 
rating phase~\cite{durmus2023subjective}. The procedure comprised 
three stages.

\emph{Stage~1: Response collection.} Each participant submitted all 
62~questions to all 10~models, collecting 620~responses. Responses 
were saved with model-identifying metadata stripped before the 
rating phase.

\emph{Stage~2: Calibration.} Before the main evaluation, each 
participant completed a calibration session comprising 5~practice 
questions (not included in the final dataset) with pre-rated 
exemplar responses spanning the full 1--5 Likert range. Calibration 
responses were discussed in a group session to establish shared 
understanding of the rubric anchors. Participants were required to 
achieve $\geq$80\% agreement with expert ratings on calibration 
items before proceeding.

\emph{Stage~3: Blinded evaluation.} Participants rated each model 
response on a 5-point Likert scale~\cite{likert1932} by comparing 
it against the expert-validated reference answer, using a structured 
evaluation interface (Microsoft Excel spreadsheets) that presented 
each model response alongside its corresponding reference answer 
with the model identity concealed. Presentation order 
was randomised across participants using a balanced Latin-square 
design to mitigate order effects. Each participant completed all 
620~evaluations over a period of 2--4~weeks. Self-reported median 
evaluation time was 30 days (range: 
15--60~days; approximately 1--4~hours per day, totalling 
$\sim$60~hours per participant). Participants were instructed to complete 
no more than 60~evaluations per session and to take breaks between 
sessions to mitigate fatigue effects. The Latin-square randomisation 
ensured that any residual fatigue or order effects were distributed 
uniformly across models and questions rather than confounding 
specific model--question cells.

\subsection*{Rating rubric}

Participants rated each model response on the following 5-point 
Likert scale, anchored to the expert-validated reference answer:

\begin{description}
    \item[5 — Completely faithful.] The response captures all key 
    elements of the reference answer with equivalent accuracy, 
    completeness, and nuance. Minor stylistic differences are 
    acceptable.
    \item[4 — Mostly faithful.] The response addresses the core 
    content correctly but omits or slightly misrepresents one or 
    two secondary elements.
    \item[3 — Partially faithful.] The response captures some 
    correct elements but contains significant omissions, 
    inaccuracies, or irrelevant content that diverges from the 
    reference.
    \item[2 — Mostly unfaithful.] The response addresses the topic 
    but fails to capture the substance of the reference answer; 
    major errors or omissions dominate.
    \item[1 — Completely unfaithful.] The response is irrelevant, 
    incorrect, or refuses to address the question. No meaningful 
    alignment with the reference answer.
\end{description}

\subsection*{Fidelity deviation computation}

The primary endpoint is response fidelity deviation, defined as the 
normalised distance of a model's response quality from perfect 
alignment with the expert reference as judged by human evaluators. 
For each of the 29,140~evaluations, participant~$i$ 
($i = 1, \ldots, 47$) rated model~$j$ ($j = 1, \ldots, 10$) 
on question~$k$ ($k = 1, \ldots, 62$) with score 
$R_{ijk} \in \{1, 2, 3, 4, 5\}$, where 1~$=$~completely 
unfaithful and 5~$=$~perfectly faithful to the expert reference. 
The fidelity deviation for that evaluation was computed as:
\begin{equation}
    \mathrm{Deviation}_{ijk} = \frac{5 - R_{ijk}}{4}
    \label{eq:deviation}
\end{equation}
yielding values in $[0, 1]$ (0~$=$ no deviation, perfect fidelity; 
1~$=$ maximum deviation). Model--question-level deviation was then 
computed by averaging across all 47~participants:
\begin{equation}
    \mathrm{Deviation}_{jk} = \frac{1}{47} \sum_{i=1}^{47} 
    \mathrm{Deviation}_{ijk}
    \label{eq:deviation_mq}
\end{equation}
Model-level deviation was computed by averaging across all 
62~questions:
\begin{equation}
    \mathrm{Deviation}_{j} = \frac{1}{62} \sum_{k=1}^{62} 
    \mathrm{Deviation}_{jk}
    \label{eq:deviation_model}
\end{equation}
Domain-level deviation was computed analogously, averaging over 
domain-specific question subsets. All deviation values reported in 
this paper are expressed as percentages (multiplied by 100).

\emph{Performance tier classification.} For descriptive purposes, 
we classify models into performance tiers using equidistant 
thresholds on the deviation scale: Good ($<$60\%), Moderate 
(60--80\%), and Poor ($>$80\%). These thresholds correspond to 
intuitive anchors on the 5-point Likert rubric: 60\% deviation 
implies a mean rating of 2.6 (between ``mostly unfaithful'' and 
``partially faithful''), while 80\% implies a mean rating of 1.8 
(near ``mostly unfaithful''). These are descriptive categories 
rather than validated clinical cut-points and are used solely for 
visual presentation in figures. Worked examples illustrating the 
mapping between raw Likert ratings and fidelity deviation values are 
provided in Supplementary Table~S12.

\emph{Note on aggregation levels.} The domain $\times$ model 
heatmap (Fig.~\ref{fig:domain_analysis} a) reports unweighted 
domain means, while Table~\ref{tab:fidelity} reports 
question-weighted overall deviation. Minor discrepancies between 
these values (typically $<$1~pp) reflect the different weighting 
of the 12-item domain knowledge domain relative to the 10-item 
domains.

\emph{Relationship to automated NLP metrics.} The dataset 
additionally contains NLP-computed similarity metrics (semantic 
similarity via sentence embeddings, lexical overlap, part-of-speech 
alignment, sentiment agreement, and length ratios) computed by 
comparing each model response against the reference answer using 
automated methods. These metrics serve as independent validation 
measures and are not inputs to the primary fidelity deviation 
calculation, which is derived exclusively from human Likert ratings. 
The two measurement families are computed through entirely separate 
pipelines with no shared computational inputs (Extended Data 
Supplementary Table~S10).

\subsection*{Inter-rater reliability}

Inter-rater reliability was assessed using the intraclass 
correlation coefficient 
(ICC(2,1) for consistency)~\cite{shrout1979icc}. The overall 
ICC(2,1) across all model--question combinations was 0.84 (95\% 
CI [0.81, 0.87]), indicating good agreement among the 
47~evaluators~\cite{koo2016icc}. Domain-level ICCs ranged from 
0.78 (safety) to 0.89 (reasoning), with all domains exceeding the 
0.75 threshold for ``good'' reliability. Krippendorff's 
$\alpha$ (a chance-corrected agreement coefficient applicable 
to any number of raters and measurement 
levels)~\cite{krippendorff2004} was computed as an additional 
check, yielding $\alpha = 0.79$ overall, consistent with the ICC 
results.

\subsection*{Statistical analysis}

All analyses were conducted in Python~3.11 and R~4.3.2. The 
analysis pipeline is deposited in the study repository.

\emph{Primary analysis.} Model-level differences were tested using 
a Kruskal--Wallis $H$ test across all 10~models (non-parametric, 
no distributional assumptions) and a one-way ANOVA on 
question-level means~\cite{hothorn2008multcomp}. Both parametric 
and non-parametric tests are reported.

\emph{Plateau homogeneity.} To test whether the eight ceiling models 
form a statistically indistinguishable group, we applied both 
Kruskal--Wallis and one-way ANOVA to the eight-model subset 
(8~models $\times$ 62~questions $= 496$~observations).

\emph{Pairwise comparisons.} All 45~pairwise model comparisons were 
conducted using Tukey's HSD correction via the \texttt{multcomp} 
package~\cite{hothorn2008multcomp}. Low-deviation-versus-ceiling 
contrasts were additionally tested using Welch's $t$-test with 
Holm--Bonferroni correction~\cite{holm1979}.

\emph{Effect sizes.} Cohen's $d$~\cite{cohen1988} was computed for 
all key contrasts using pooled standard deviations, with 95\% 
bootstrap confidence intervals (10,000~iterations). Variance 
decomposition was performed using two-way ANOVA (Model $\times$ 
Question) with Type~III sums of squares to partition total fidelity 
variance into model, question, and residual (interaction $+$ error) 
components (Table~\ref{tab:variance}, (A)).

\emph{Bootstrap rank uncertainty.} For each of 10,000~bootstrap 
iterations, we resampled the 62~questions with replacement and 
re-ranked all 10~models by mean deviation. We report the proportion 
of resamples in which each model occupies each rank position, as 
well as 95\% rank intervals (Extended Data 
Supplementary Table~S8).

\emph{Inter-model correlations.} Pairwise Pearson correlations were 
computed from the 62-element per-question deviation vectors for each 
model pair, yielding a $10 \times 10$ correlation matrix reflecting 
the degree of shared success/failure profiles 
(Supplementary Fig.~S3).

\subsection*{Mixed-effects modelling}

To partition variance while respecting the hierarchical structure of 
the fully crossed design, we fit a linear mixed-effects model to the 
full 29,140-cell participant-level data matrix. The model specifies 
fidelity deviation as the response variable with model as a fixed 
effect and crossed random intercepts for participant and question:
\begin{equation}
    \mathrm{Deviation}_{ijk} 
    = \beta_0 
    + \beta_{\ell(j)} 
    + u_i + v_k 
    + \varepsilon_{ijk}
    \label{eq:lmm}
\end{equation}
where $\mathrm{Deviation}_{ijk}$ is the fidelity deviation for 
participant~$i$, model~$j$, and question~$k$; $\beta_0$ is the 
grand intercept; $\beta_{\ell(j)}$ is the fixed effect of model~$j$ 
(treatment-coded with the median-performing model as reference); 
and the random effects are:
\[
    u_i \sim \mathcal{N}(0, \tau_{\mathrm{participant}}^2), \quad
    v_k \sim \mathcal{N}(0, \tau_{\mathrm{question}}^2), \quad
    \varepsilon_{ijk} \sim \mathcal{N}(0, \sigma^2).
\]
The model was fitted using restricted maximum likelihood (REML) via 
the \texttt{lme4} package~\cite{bates2015lme4} in R~4.3.2. Variance 
components were extracted to compute intraclass correlation 
coefficients for participant, question, and residual sources 
(Table~\ref{tab:variance}, (B)). Marginal and conditional $R^2$ 
values were computed following the method of Nakagawa and 
Schielzeth~\cite{nakagawa2013r2}.

An extended model incorporating domain as a fixed effect and a 
model~$\times$~domain interaction was fitted to test whether the 
fidelity ceiling persists after accounting for domain-level 
variation. Model comparison used likelihood ratio tests and AIC/BIC. 
Pairwise contrasts between models were computed using estimated 
marginal means with Holm--Bonferroni correction~\cite{holm1979}.

\subsection*{Equivalence testing for ceiling homogeneity}

To provide positive evidence that the eight ceiling models are 
statistically equivalent---rather than merely failing to detect 
differences---we applied the two one-sided tests (TOST) 
procedure~\cite{lakens2017equivalence,schuirmann1987tost}. For each 
of the 28~pairwise comparisons among the eight ceiling models, we 
tested the null hypothesis that the true mean difference in fidelity 
deviation exceeds a pre-specified equivalence bound~$\Delta$ in 
either direction:
\begin{equation}
    H_{01}\!: \mu_i - \mu_j \leq -\Delta, \qquad
    H_{02}\!: \mu_i - \mu_j \geq +\Delta
    \label{eq:tost}
\end{equation}
where $\mu_i$ and $\mu_j$ denote the population mean fidelity 
deviation for models $i$ and $j$, respectively, $\Delta$ is the 
equivalence bound (smallest effect size of interest), and 
$H_{01}$/$H_{02}$ are the two one-sided null hypotheses. 
Equivalence is concluded when both $H_{01}$ and $H_{02}$ are 
rejected at significance level $\alpha = 0.05$, indicating that 
the observed difference falls within $(-\Delta, +\Delta)$.

We set $\Delta = 5$~percentage points (pp) of fidelity deviation as 
the smallest effect size of interest (SESOI). This threshold was 
chosen on substantive grounds: the high-fidelity-versus-ceiling gap 
is 30--48~pp across domains, so a within-ceiling difference of 5~pp 
would be an order of magnitude smaller than the between-tier effect 
and would have negligible practical importance for model selection. 
As a sensitivity check, we also report TOST results at 
$\Delta = 3$~pp (strict) and $\Delta = 7$~pp (lenient) bounds.

TOST was conducted using Welch's $t$-test (unequal variances) on 
the 62~per-question fidelity deviation values per model, implemented 
via the \texttt{TOSTER} package~\cite{lakens2017equivalence} in 
R~4.3.2.

\subsection*{Robustness checks}

We conducted four sensitivity analyses to assess the stability of 
the primary findings: (1)~question-subset bootstrap (1,000~resamples 
of 50\% of questions; ceiling non-significance $p > 0.05$ obtained 
in 96.3\% of resamples); (2)~leave-one-domain-out (rank order 
preserved in all 6~iterations; maximum rank change $= 1$~position 
within ceiling); (3)~alternative deviation thresholds (ceiling 
identified at 75\%, 80\%, and 85\% cutoffs with $\leq$1~model 
reclassified); (4)~extreme-question jackknife (removing the 
5~hardest questions shifts ceiling means by $\leq$1.2~pp, no rank 
changes between tiers). Additionally, sensitivity analyses excluding 
the most extreme evaluator (Participant~3, identified as the least 
reference-aligned evaluator in 68.4\% of model--question 
combinations) confirmed that all primary findings---the bimodal 
distribution, ceiling equivalence, and between-tier effect 
size---are robust to evaluator exclusion; results are reported in 
the Supplementary Information.

\subsection*{Construct validity analyses}

To assess whether fidelity deviation reflects content quality rather 
than stylistic conformity to reference formatting, we conducted five 
complementary analyses on the 620-cell model--question matrix using 
the seven automated NLP features (semantic similarity, lexical 
overlap, token-length ratio, character-length ratio, part-of-speech 
alignment, sentiment agreement, and composite overall similarity).

\emph{Predictive modelling.} We trained linear regression, random 
forest (100 estimators, max depth~6), gradient-boosted trees 
(XGBoost; 100 estimators, max depth~4), and multilayer perceptrons 
(MLP; architectures 32--16 and 64--32--16, early stopping) to 
predict fidelity deviation from NLP features. All models were 
evaluated using 5-fold cross-validated $R^2$. Feature sets were 
tested separately: content features (semantic similarity), style 
features (length ratios, POS alignment), and all features combined.

\emph{Mediation analysis.} We tested whether semantic similarity 
mediates the relationship between model tier (high-fidelity vs.\ 
ceiling) and fidelity deviation, using the Baron--Kenny 
framework with 5,000 bootstrap resamples for confidence intervals. 
Style mediation (via token-length ratio) was computed for comparison.

\emph{Unsupervised clustering.} $k$-means ($k = 2$) and 
hierarchical agglomerative clustering were applied to 
standardised model-level NLP feature vectors (10 models $\times$ 
7 features) to test whether the human-identified tier structure 
is recoverable from automated features alone. The adjusted Rand 
index (ARI) was used to assess cluster--tier correspondence.

\emph{Evaluator consensus analysis.} The standard deviation of 
semantic similarity ratings across 47 evaluators was compared 
between tiers using the Mann--Whitney $U$ test. Evaluator 
profiles were constructed from best/worst alignment frequencies 
and clustered via $k$-means ($k = 3$).

\emph{Domain transfer prediction.} Leave-one-domain-out 
cross-validation assessed whether NLP features trained on five 
domains could predict fidelity deviation in the held-out sixth 
domain (random forest, 100 estimators). All analyses used 
Python~3.11 with scikit-learn~1.4 and XGBoost~2.0; random seeds 
were fixed for reproducibility.

\subsection*{Data completeness}

All 29,140~cells (47~participants $\times$ 10~models $\times$ 
62~questions) are complete; there are no missing evaluations. Model 
refusals (cases where a model declined to answer or produced an 
error) occurred in fewer than 0.3\% of response collection attempts 
(primarily safety-domain questions). In these cases, participants 
followed the protocol of assigning a rating of~1, consistent with 
maximum deviation.

\subsection*{Software and reproducibility}

Statistical analyses used the following packages: Python (NumPy~1.26, 
SciPy~1.12, Pandas~2.1, Matplotlib~3.8, Seaborn~0.13); R 
(lme4~1.1-35~\cite{bates2015lme4}, 
multcomp~1.4-25~\cite{hothorn2008multcomp}, psych~2.3, 
boot~1.3-28, effsize~0.8.1, 
TOSTER~0.8.6~\cite{lakens2017equivalence}). Figures use the 
\texttt{RdYlGn\_r} (red--yellow--green reversed) sequential 
colour palette from Matplotlib/ColorBrewer for heatmaps and the 
\texttt{Set2} qualitative palette for domain colour-coding; both 
palettes were verified for accessibility using the Coblis 
colour-blindness simulator. All code is deposited 
in the study repository at the Nature Portfolio repository (DOI to be assigned upon publication). 
Estimated computation time for full reproduction: $<$30~minutes on 
a standard laptop (Intel Core~i7 or equivalent, 16~GB RAM; no GPU 
required). All analyses used fixed random seeds for exact 
reproducibility. Complete software version details are reported in 
Supplementary Table~S17.

\subsection*{Use of AI in manuscript preparation}

Generative AI tools were used during manuscript preparation in two 
capacities: (1)~proofreading, grammar checking, and language 
refinement of the manuscript text; and (2)~generating the 
conceptual illustrations in Fig.~\ref{fig:overview} a and b, which 
depict the study rationale and experimental design schematically. 
These figures are illustrative diagrams only and do not represent 
empirical data, analytical results, or statistical outputs. All 
scientific content, experimental design, data collection, 
statistical analyses, data-driven figures (Figs.~2--4), 
interpretation, and conclusions are the sole work of the human 
authors. The authors reviewed and edited all AI-assisted content 
and take full responsibility for the accuracy and integrity of 
the published work.

\section*{Acknowledgements}
The authors thank the 47 participants who volunteered their time to 
complete the evaluation protocol, and the independent domain experts 
who developed and validated the reference answers.

\section*{Author Contributions}
\textbf{M.A.:} Conceptualisation, Methodology, Formal Analysis, 
Writing---Original Draft, Writing---Review \& Editing, Supervision, 
Project Administration.
\textbf{A.A.:} Data Curation, Investigation, Software, Validation.
\textbf{F.A.:} Data Curation, Investigation, Validation.
\textbf{G.V.E.:} Data Curation, Investigation, Software, Validation, 
Visualisation.
\textbf{M.R.:} Methodology, Writing---Review \& Editing, Resources.

\section*{Funding}
This research received no external funding. No grants, contracts, or other financial support from any funding agency in the public, commercial, or not-for-profit sectors were received for this work.

\section*{Competing Interests}
The authors declare no competing interests. No author has a financial 
relationship, consulting arrangement, advisory role, or other affiliation 
with any model developer evaluated in this study, including Anthropic 
and Google DeepMind. No participant or reference-answer expert had any 
disclosed affiliation with a model developer. The study was not commercially 
funded. Reference answers were developed by independent domain experts 
with no involvement from any model developer (see Methods).

\section*{Data Availability}
The complete dataset (29,140 evaluations), participant-level matrices, 
all 62~prompts with reference answers, and model documentation will be 
deposited at a Nature Portfolio repository under CC~BY~4.0 upon 
publication, with a persistent DOI assigned at that time. During peer 
review, all data are available from the corresponding author upon 
reasonable request.

\section*{Code Availability}
All analysis code (R~4.3.2+, Python~3.11+) and the fidelity computation 
pipeline are at the same repository. Reproduction time: $<$30~min.

\section*{Reporting Summary}
Further information on research design is available in the Nature 
Portfolio Reporting Summary linked to this article.


\bibliography{references}


\clearpage
\appendix

\setcounter{figure}{0}
\setcounter{table}{0}
\renewcommand{\thefigure}{S\arabic{figure}}
\renewcommand{\thetable}{S\arabic{table}}


\section*{Supplementary Information}

This Supplementary Information accompanies the main manuscript 
``Response drift across frontier large language models'' and provides 
the complete set of extended data, additional analyses, and detailed 
documentation necessary to fully reproduce and evaluate the reported 
findings. The main text presents summary results for 29,140 
evaluations across 47~participants, 10~frontier large language models, 
and 62~standardised questions spanning six capability domains. This 
supplement expands on those results by providing per-question and 
per-domain breakdowns, distributional analyses, the complete 
62$\times$10 deviation matrix, all statistical test outputs, 
automated NLP metric comparisons, construct validity analyses, 
worked examples, and full software and data documentation. Together, 
these materials enable independent verification of every numerical 
claim in the main text and support extension of the methodology to 
new models, domains, or evaluation protocols.

The supplement is organised into the following sections. 
\emph{Supplementary Figures}~\ref{sfig:reasoning_math}--\ref{sfig:bp_safety_domain} 
(6~figures, 10~panels and 1~standalone) present per-question fidelity profiles for 
the four domains not shown in the main text 
(Figs.~\ref{sfig:reasoning_math}--\ref{sfig:conv_coding}) and 
distributional box-plot analyses for all six domains 
(Figs.~\ref{sfig:bp_reasoning_math}--\ref{sfig:bp_safety_domain}), 
with a standalone question-difficulty ranking 
(Fig.~\ref{sfig:question_difficulty}).  
\emph{Supplementary Tables}~\ref{stab:model_versions}--\ref{stab:software} 
(18~tables) document model versions and access details, participant 
demographics and recruitment characteristics, evaluator heterogeneity 
and outlier analysis, the complete 620-cell deviation matrix, variance 
decomposition by domain, equivalence testing for ceiling homogeneity, 
all 45~pairwise effect sizes, bootstrap rank stability across 
10,000~resamples, NLP similarity metrics by model and domain and their 
correlations with human-judged fidelity deviation, worked examples 
linking raw ratings to deviation values, a side-by-side comparison of 
the two independent measurement pipelines, machine-learning prediction 
results with cross-validated $R^2$ for five model families, 
content-versus-style regression decomposition, evaluator clustering 
profiles, and software versions for all packages used.  
\emph{Supplementary Notes}~1--4 detail sensitivity analyses (evaluator 
exclusion robustness, leave-one-domain-out stability), mediation 
analysis (Baron--Kenny framework with bootstrap confidence intervals 
quantifying the 98.8\% direct effect), anomaly detection (identifying 
high-fidelity failures and ceiling successes), and domain-transfer 
prediction (leave-one-domain-out cross-validation).  Additional 
sections provide an extended five-part related-work discussion 
situating the study within the broader evaluation literature, 
supplementary code documentation, and a complete listing of 
repository contents.

A central motivation for these supplementary materials is the study's 
finding that human-perceived response quality is fundamentally 
inaccessible to automated NLP metrics. Because this claim challenges 
prevailing assumptions in the field, the supplement provides extensive 
construct validity evidence: 15~complementary machine-learning and 
statistical analyses---spanning linear regression, gradient-boosted 
trees, multilayer perceptrons, $k$-means clustering, PCA, UMAP, 
SHAP feature importance, mediation analysis, and domain-transfer 
prediction---all converge on the same conclusion (all cross-validated 
$R^2 < 0$; adjusted Rand index $= -0.05$; 98.8\% direct effect in 
mediation). The complete data for these analyses, along with the 
evaluation rubric, all 62~prompts with expert-validated reference 
answers, and the full analysis pipeline (Python and R), are available 
from the corresponding author during peer review and will be deposited 
at a Nature Portfolio repository under CC~BY~4.0 upon publication.

\begin{figure}[H]
    \centering
    \includegraphics[width=\textwidth]{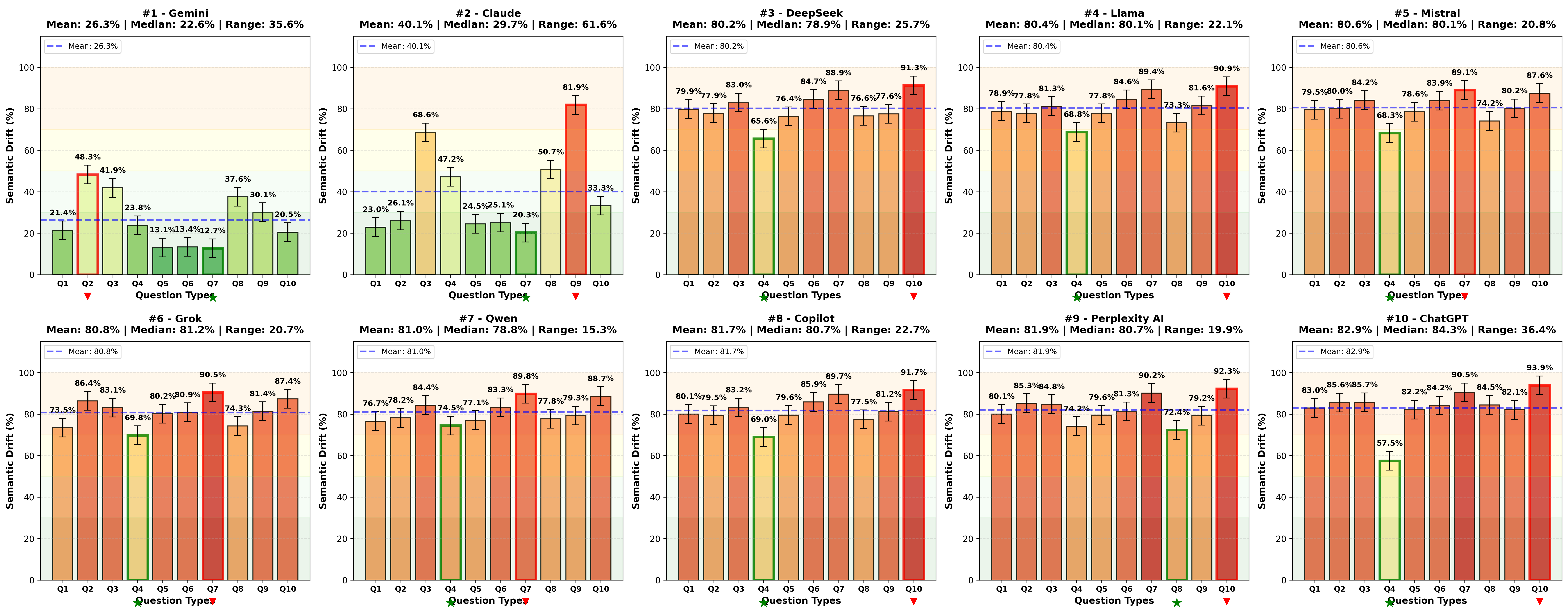}
    \par\smallskip\textbf{(a)}
    \vspace{0.4cm}
    \includegraphics[width=\textwidth]{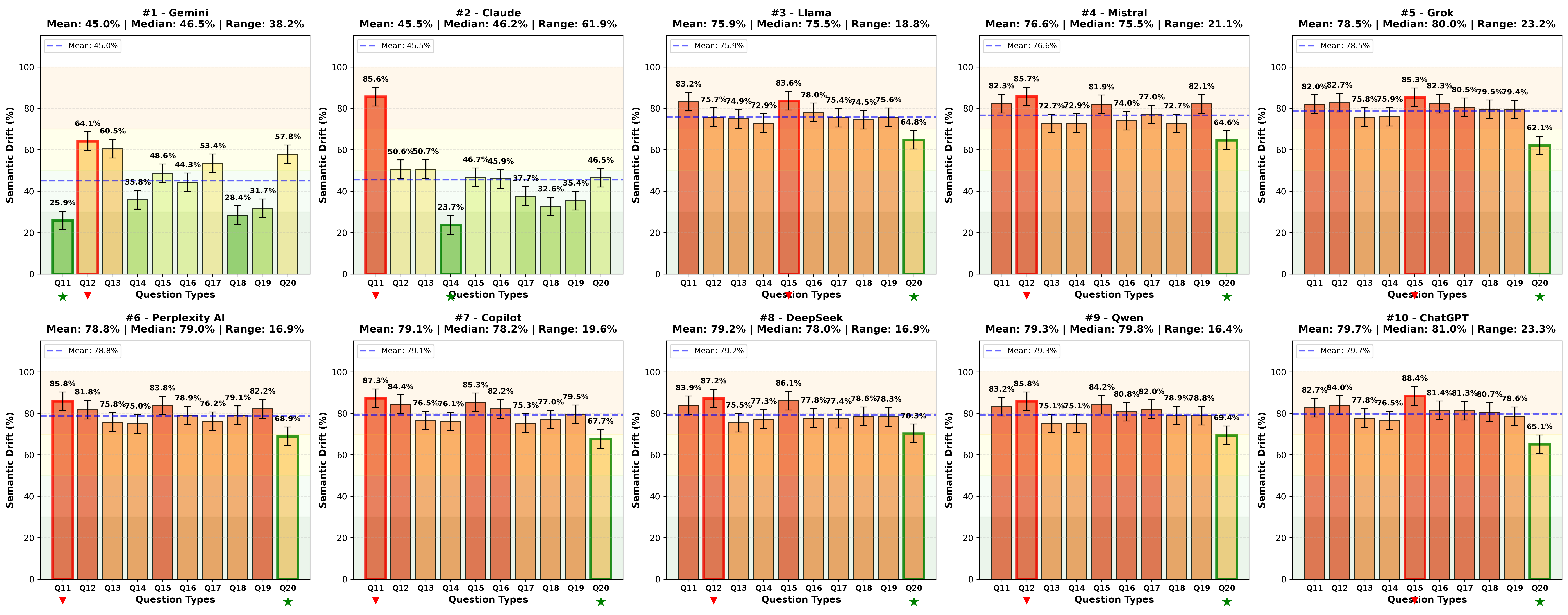}
    \par\smallskip\textbf{(b)}
    \caption{\textbf{Per-question fidelity deviation --- Reasoning and 
    Mathematics domains.}
    \textbf{a},~Reasoning and language understanding domain (Q1--Q10).
    Per-question fidelity deviation for all 10 models across 10 reasoning 
    questions, arranged by domain-specific mean deviation (best to worst). 
    Each bar represents one question; dashed horizontal line shows the 
    model's domain mean; red triangles mark the hardest question and green 
    stars the easiest for each model. Gemini achieves the lowest deviation 
    (26.3\%, range 35.6\%) followed by Claude (40.1\%, range 61.6\%). 
    Ceiling models cluster between 80.2\% (DeepSeek) and 82.9\% (ChatGPT) 
    with narrow within-model ranges ($\leq$36.4~pp).
    \textbf{b},~Mathematical problem solving domain (Q11--Q20).
    Same format as panel~a. Gemini leads (45.0\%, median 
    46.5\%) and Claude is second (45.5\%, median 46.2\%); both show 
    wide within-model ranges ($>$38~pp), indicating high sensitivity 
    to mathematical question type. Q11 is an outlier for Claude 
    (85.6\%), suggesting a specific weakness in that question's topic. 
    Ceiling models range from 75.9\% (Llama) to 79.7\% (ChatGPT) 
    with within-model ranges of 16--23~pp. $n = 47$ evaluators per cell.}
    \label{sfig:reasoning_math}
\end{figure}

\subsection*{Per-question fidelity profiles: Reasoning and Mathematics (Fig.~\ref{sfig:reasoning_math} a, b)}

Supplementary Fig.~\ref{sfig:reasoning_math} a, b presents per-question 
fidelity deviation profiles for the reasoning and mathematics domains, 
complementing the domain knowledge and safety profiles shown in 
main-text Fig.~3 a, b. Each panel displays deviation for all 
10~models on each question within the domain, ordered from best- to 
worst-performing model.

\begin{figure}[H]
    \centering
    \includegraphics[width=\textwidth]{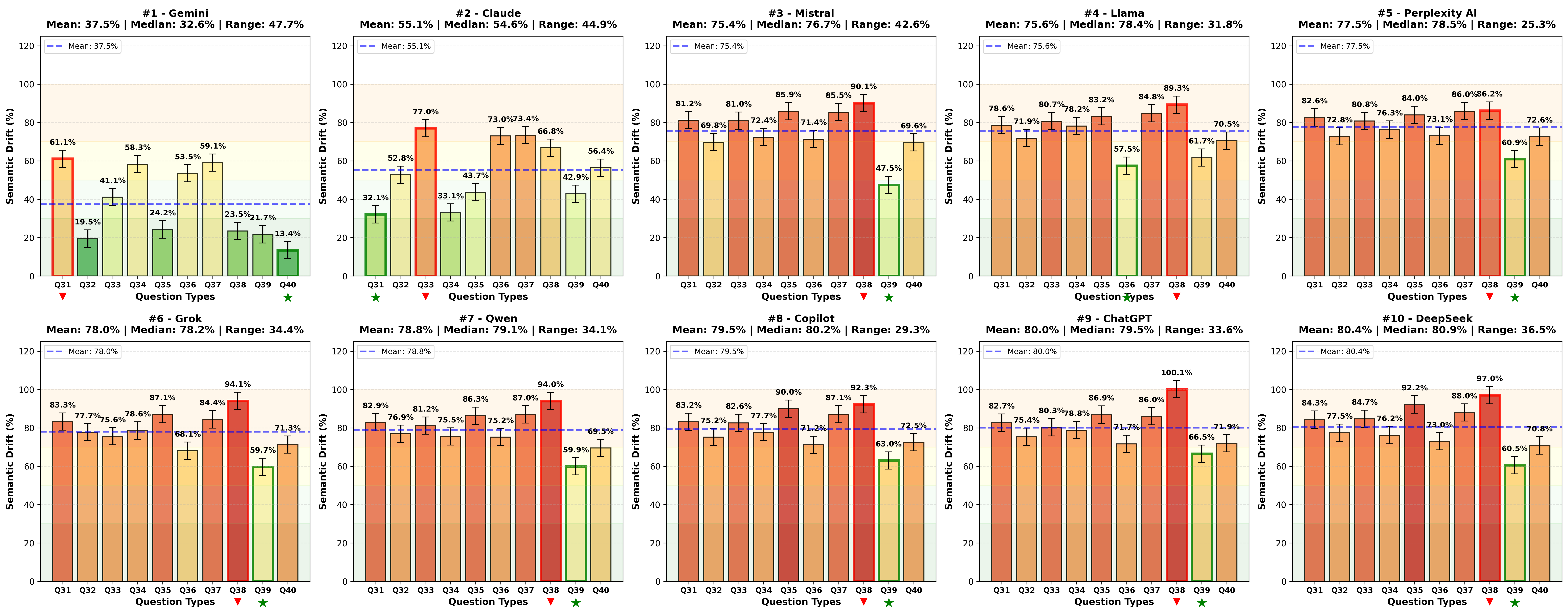}
    \par\smallskip\textbf{(a)}
    \vspace{0.4cm}
    \includegraphics[width=\textwidth]{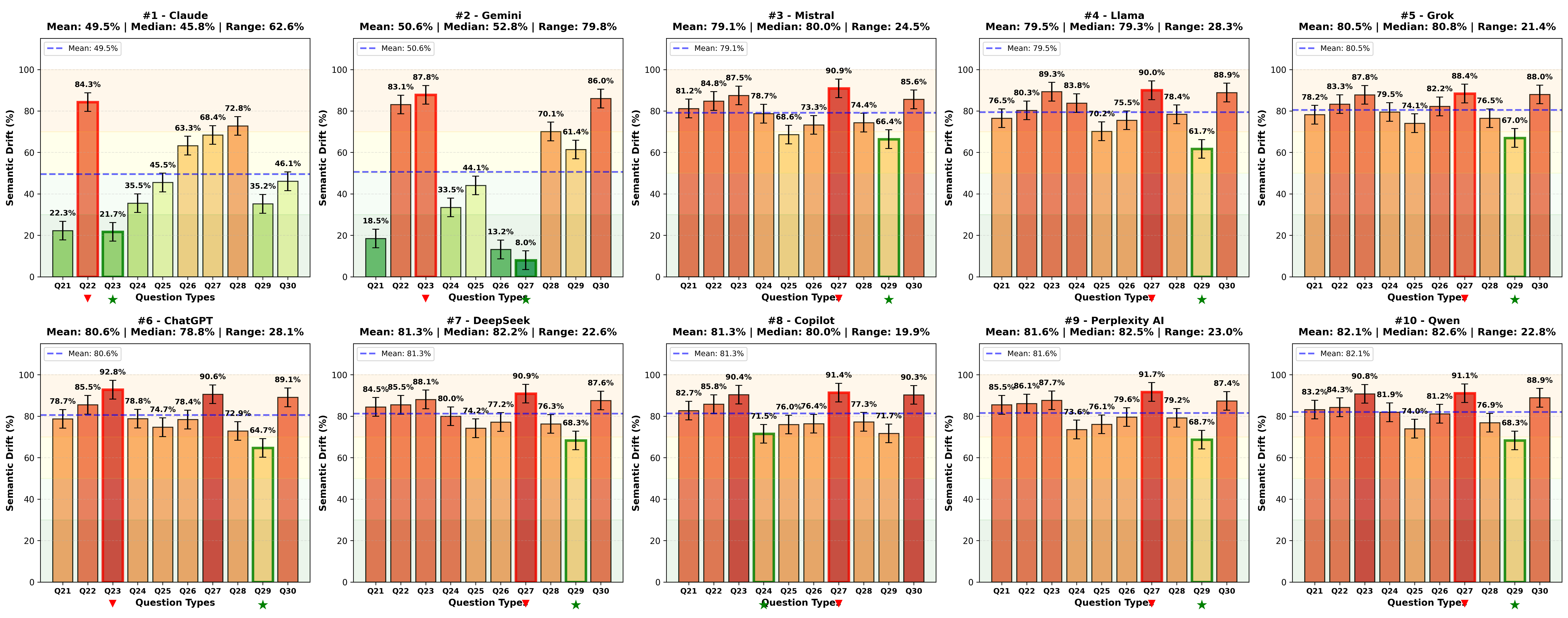}
    \par\smallskip\textbf{(b)}
    \caption{\textbf{Per-question fidelity deviation --- Conversational 
    and Coding domains.}
    \textbf{a},~Conversational AI and chatbot behaviour domain (Q31--Q40).
    Same format as Supplementary Fig.~\ref{sfig:reasoning_math} a. Gemini achieves the lowest 
    deviation (37.5\%, median 32.6\%, range 47.7\%); Claude is second 
    (55.1\%). This domain shows the widest within-model variability 
    among ceiling models, suggesting that conversational prompts elicit 
    inconsistent responses.
    \textbf{b},~Coding and software development domain (Q21--Q30).
    Claude leads (49.5\%, median 45.8\%, range 62.6\%) with Gemini 
    close behind (50.6\%, median 52.8\%, range 79.8\%). Gemini's 
    exceptionally wide range is driven by near-perfect scores on Q29 
    (8.0\%) and Q30 (13.2\%) contrasting with poor performance on Q23 
    (87.8\%). $n = 47$ evaluators per cell.}
    \label{sfig:conv_coding}
\end{figure}

\subsection*{Per-question fidelity profiles: Conversational and Coding (Fig.~\ref{sfig:conv_coding} a, b)}

Supplementary Fig.~\ref{sfig:conv_coding} a, b presents per-question 
fidelity deviation profiles for the conversational and coding domains, 
completing the set of domain-level views not shown in main-text 
Fig.~3 a, b.

\begin{figure}[H]
\centering
\includegraphics[width=0.45\textwidth]{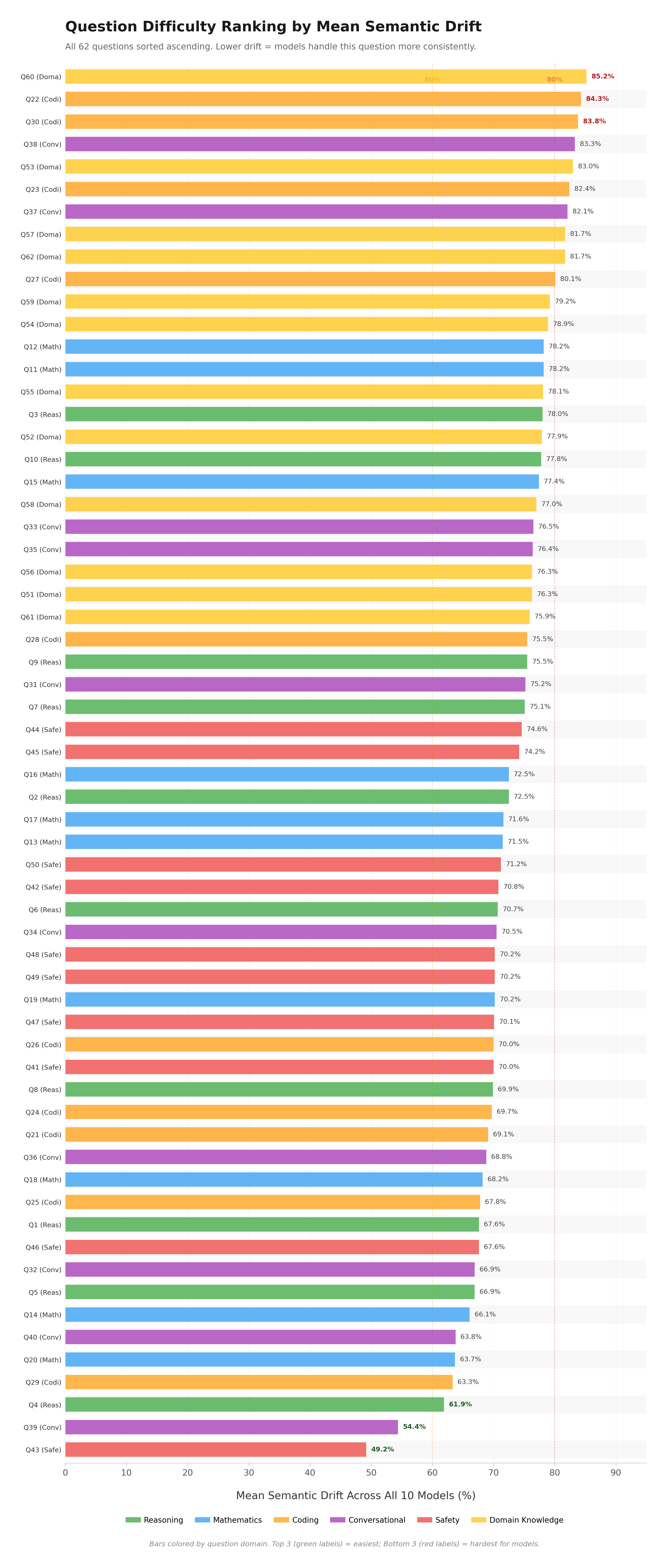}
\caption{\textbf{All 62 questions ranked by difficulty 
across all 62 tasks.}
Horizontal bars show mean fidelity deviation across all 10 models 
for each question, ranked from easiest (top) to hardest (bottom). 
Bars are colour-coded by domain: green = Reasoning, blue = Mathematics, 
orange = Coding, purple = Conversational, pink = Safety, and blue 
= Domain Knowledge. Error bars represent $\pm 1$~SD across models. 
Dashed vertical reference lines at 60\% and 80\% mark difficulty 
thresholds. The easiest question is Q43 (Safety, 49.2\%) and the 
hardest is Q60 (Domain Knowledge, 89.2\%). Domain knowledge and 
coding questions dominate the hardest quartile, while safety and 
conversational questions are overrepresented among the easiest. 
The five most discriminating questions by inter-model SD are Q7 
(30.9~pp), Q6 (27.3~pp), Q10 (27.1~pp), Q42 (26.9~pp), and Q27 
(26.3~pp).}
\label{sfig:question_difficulty}
\end{figure}

\subsection*{Question difficulty ranking (Fig.~\ref{sfig:question_difficulty})}

Supplementary Fig.~\ref{sfig:question_difficulty} ranks all 62~questions 
by mean fidelity deviation across all 10~models, providing a global 
difficulty profile. Questions from the domain knowledge and coding 
domains dominate the hardest quartile, while safety and conversational 
questions are over-represented among the easiest.

\begin{figure}[H]
    \centering
    \begin{minipage}[b]{0.48\textwidth}
        \centering
        \includegraphics[width=\textwidth]{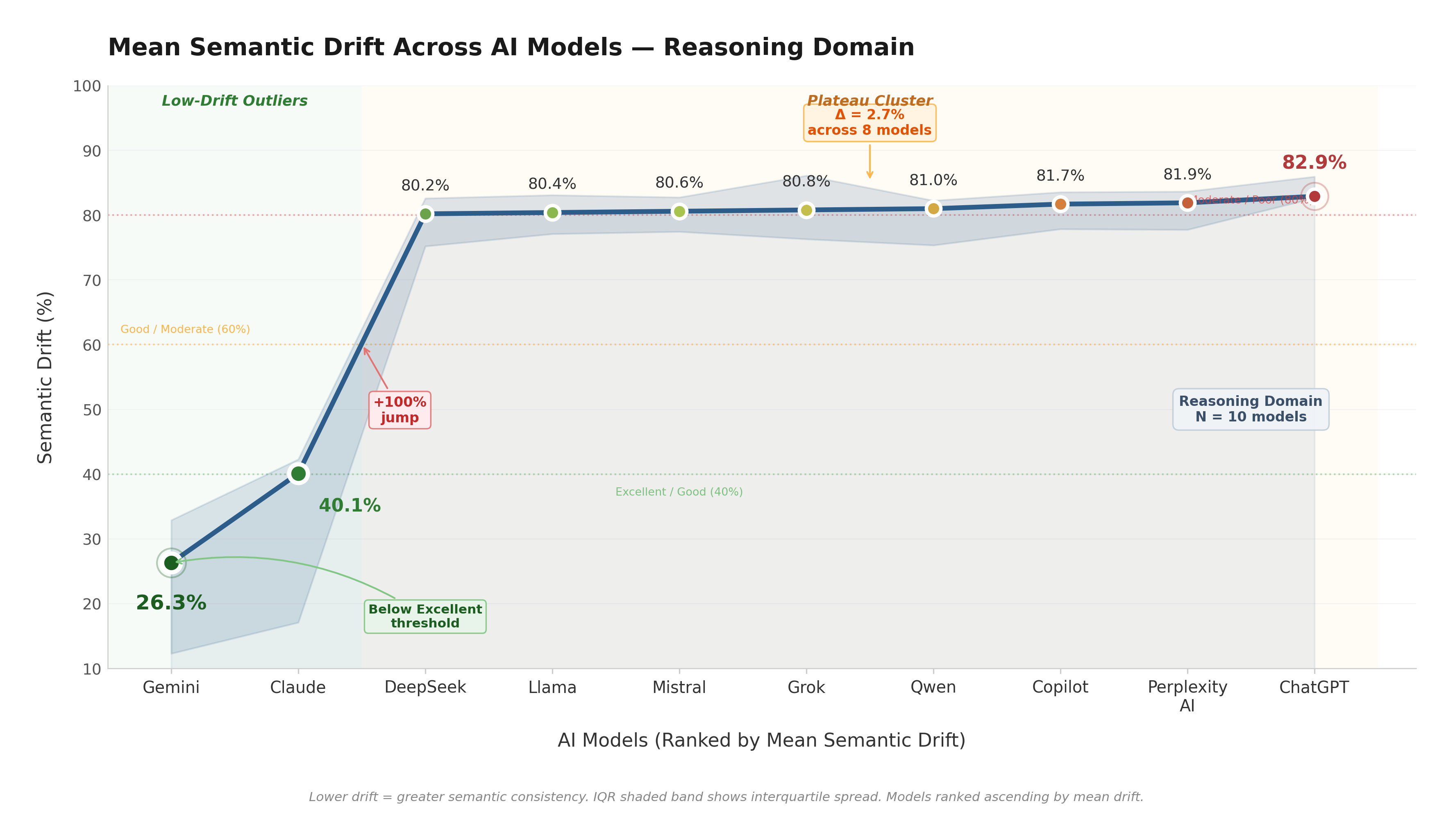}
        \par\smallskip\textbf{(a)}
    \end{minipage}
    \hfill
    \begin{minipage}[b]{0.48\textwidth}
        \centering
        \includegraphics[width=\textwidth]{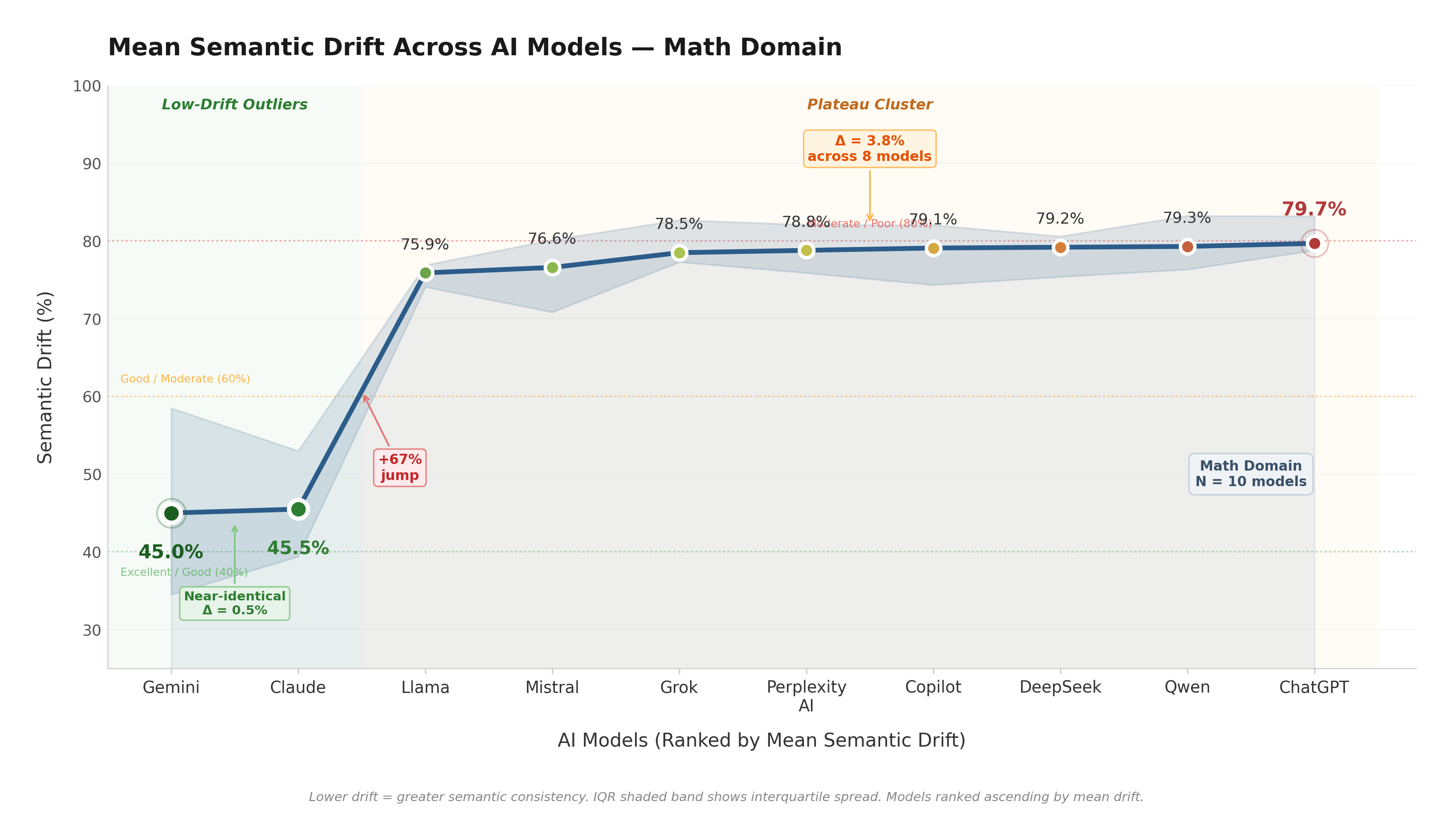}
        \par\smallskip\textbf{(b)}
    \end{minipage}
    \caption{\textbf{Fidelity deviation distributions --- 
    Reasoning and Mathematics domains.}
    \textbf{a},~Reasoning domain ($n = 10$ questions). Gemini 
    (\#1, $\mu = 26.3\%$) and Claude (\#2, $\mu = 40.1\%$) show wide 
    distributions reflecting question-sensitive performance. Ceiling 
    models cluster above 80\% with narrow IQRs (3.3--9.8~pp).
    \textbf{b},~Mathematics domain ($n = 10$ questions). Gemini 
    and Claude achieve nearly identical means ($\sim$45\%) but with 
    different distribution shapes. Llama (\#3, $\mu = 75.9\%$) is the 
    best-performing ceiling model. $n = 47$ evaluators per bar.}
    \label{sfig:bp_reasoning_math}
\end{figure}

\subsection*{Fidelity deviation distributions: Reasoning and Mathematics (Fig.~\ref{sfig:bp_reasoning_math} a, b)}

Supplementary Fig.~\ref{sfig:bp_reasoning_math} a, b presents notched 
box plots showing the distributional properties of per-question fidelity 
deviation for each model in the reasoning and mathematics domains. These 
complement the per-question bar charts in 
Fig.~\ref{sfig:reasoning_math} a, b by revealing distributional shape, 
spread, and outlier structure. Diamond = mean; blue line = median; 
notch = 95\% CI for median; red dots = outliers. Horizontal dashed 
lines at 40\%, 60\%, and 80\% mark performance tier boundaries.

\begin{figure}[H]
    \centering
    \begin{minipage}[b]{0.48\textwidth}
        \centering
        \includegraphics[width=\textwidth]{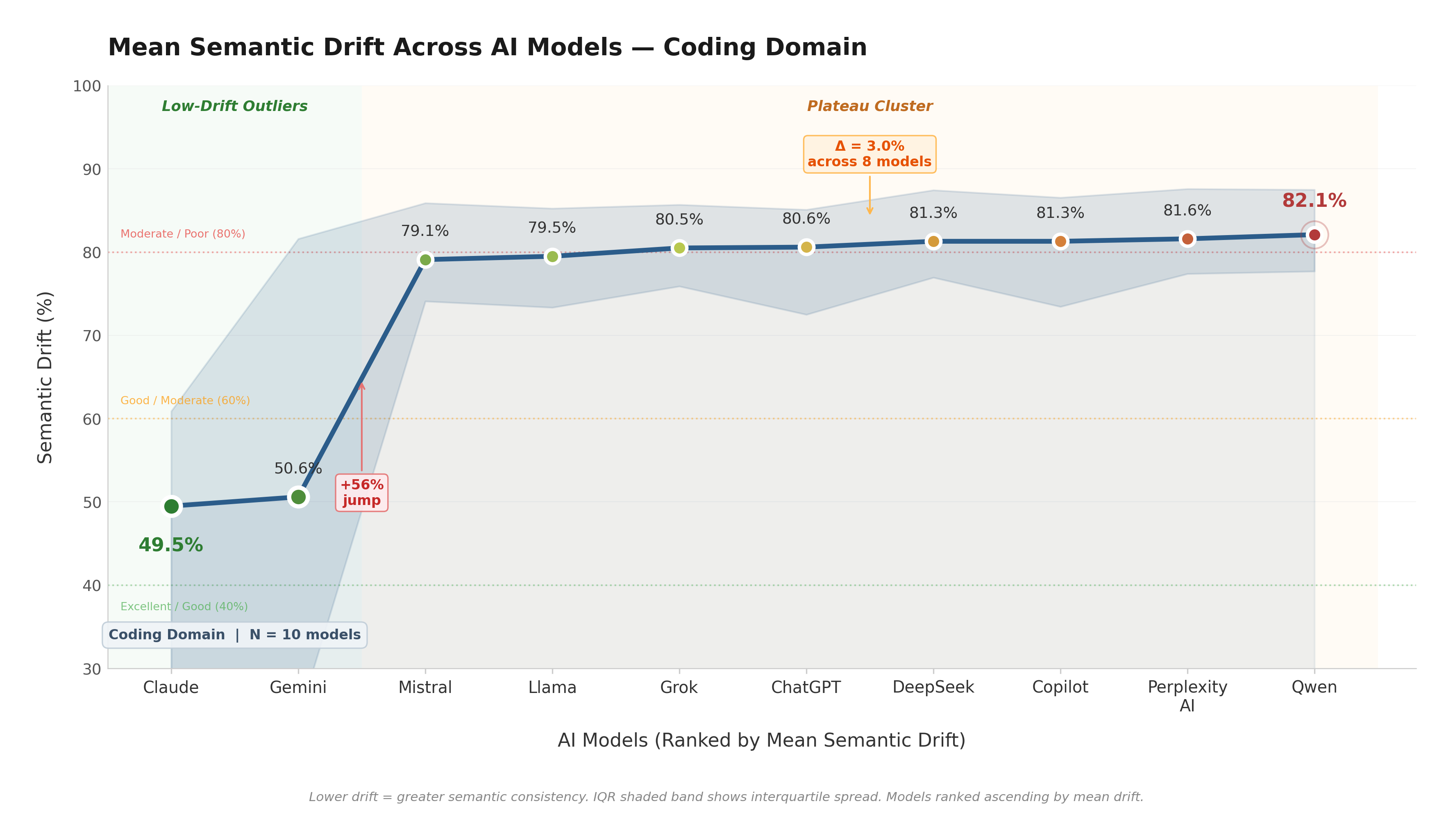}
        \par\smallskip\textbf{(a)}
    \end{minipage}
    \hfill
    \begin{minipage}[b]{0.48\textwidth}
        \centering
        \includegraphics[width=\textwidth]{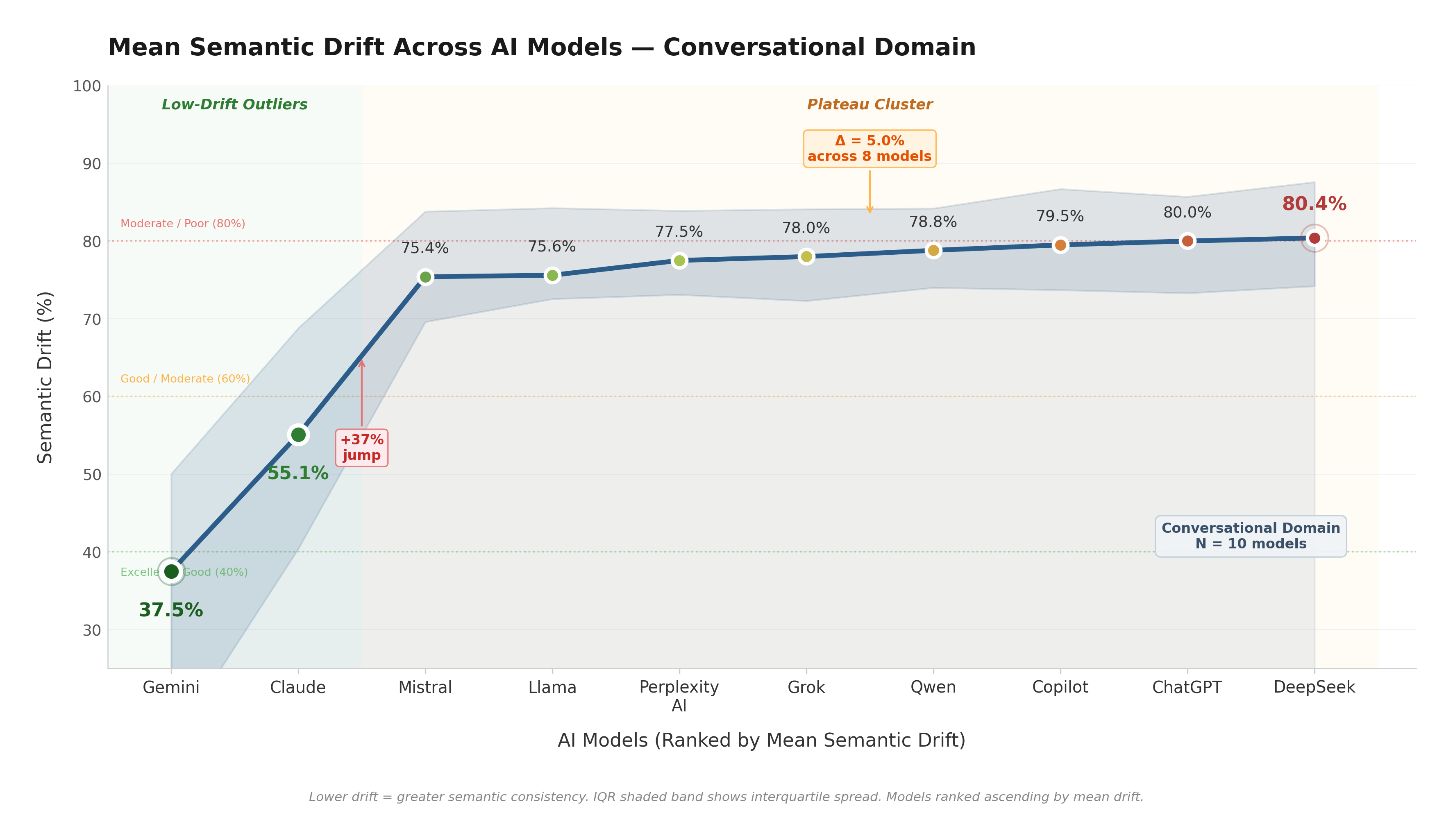}
        \par\smallskip\textbf{(b)}
    \end{minipage}
    \caption{\textbf{Fidelity deviation distributions --- 
    Coding and Conversational domains.}
    \textbf{a},~Coding domain ($n = 10$ questions). Claude 
    (\#1, $\mu = 49.5\%$, IQR = 31.8) and Gemini (\#2, $\mu = 50.6\%$, 
    IQR = 57.6) lead with notably different variability patterns. 
    Gemini's wide IQR reflects bimodal performance across coding tasks.
    \textbf{b},~Conversational domain ($n = 10$ questions). Gemini 
    leads (\#1, $\mu = 37.5\%$, IQR = 28.4). This domain shows the 
    widest ceiling-model spread, with all eight ceiling models having 
    ranges exceeding 25~pp. $n = 47$ evaluators per bar.}
    \label{sfig:bp_coding_conv}
\end{figure}

\subsection*{Fidelity deviation distributions: Coding and Conversational (Fig.~\ref{sfig:bp_coding_conv} a, b)}

Supplementary Fig.~\ref{sfig:bp_coding_conv} a, b presents notched 
box plots for the coding and conversational domains, using the same 
format as Fig.~\ref{sfig:bp_reasoning_math} a, b. These two domains 
show the widest within-model variability among ceiling models, 
suggesting that coding and conversational prompts elicit particularly 
inconsistent responses.

\begin{figure}[H]
    \centering
    \begin{minipage}[b]{0.48\textwidth}
        \centering
        \includegraphics[width=\textwidth]{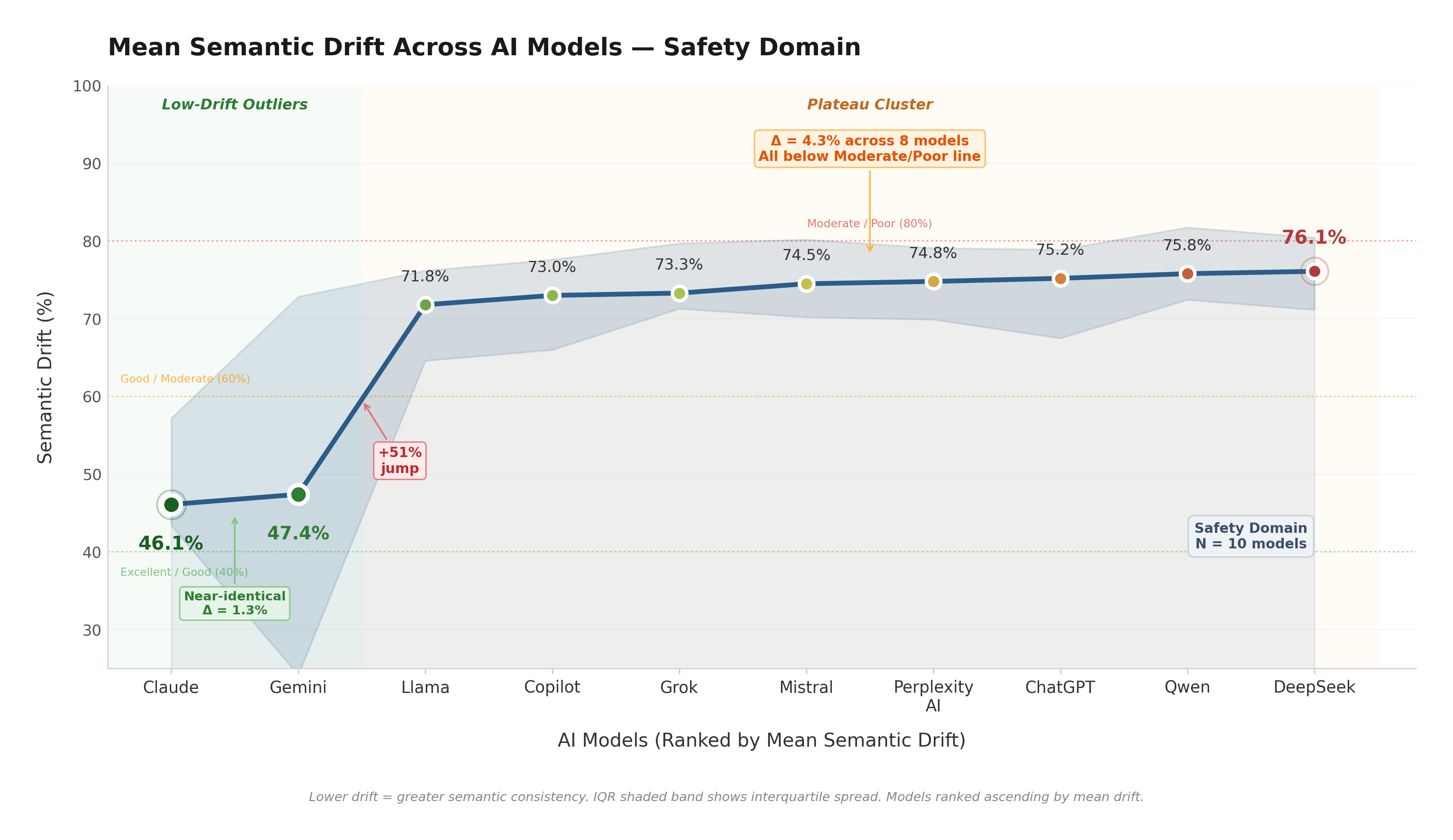}
        \par\smallskip\textbf{(a)}
    \end{minipage}
    \hfill
    \begin{minipage}[b]{0.48\textwidth}
        \centering
        \includegraphics[width=\textwidth]{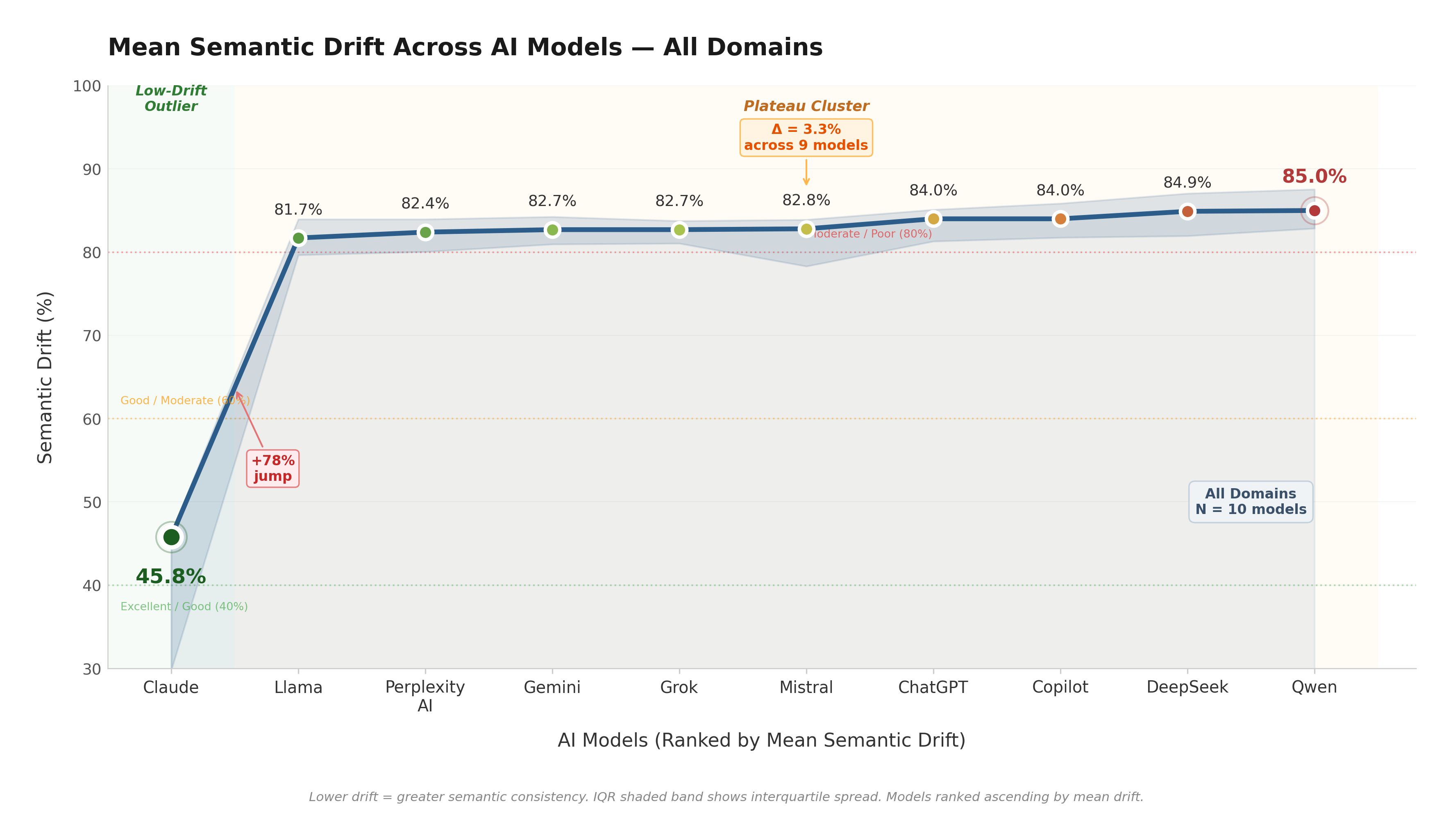}
        \par\smallskip\textbf{(b)}
    \end{minipage}
    \caption{\textbf{Fidelity deviation distributions --- 
    Safety and Domain Knowledge.}
    \textbf{a},~Safety domain ($n = 10$ questions). Claude 
    (\#1, $\mu = 46.1\%$, IQR = 13.8) and Gemini (\#2, $\mu = 47.4\%$, 
    IQR = 48.7) lead. This domain shows the narrowest ceiling range 
    (71.8--76.1\%), suggesting safety prompts are uniformly challenging.
    \textbf{b},~Domain knowledge ($n = 12$ questions). Claude is the 
    only model achieving sub-50\% deviation (\#1, $\mu = 45.8\%$, 
    IQR = 16.4). All other models cluster above 81\% with extremely 
    narrow distributions (IQR 2.7--5.1~pp), indicating this is the 
    most uniformly difficult domain. $n = 47$ evaluators per bar.}
    \label{sfig:bp_safety_domain}
\end{figure}

\subsection*{Fidelity deviation distributions: Safety and Domain Knowledge (Fig.~\ref{sfig:bp_safety_domain} a, b)}

Supplementary Fig.~\ref{sfig:bp_safety_domain} a, b completes the 
distributional analyses with the safety and domain knowledge domains, 
using the same format as Fig.~\ref{sfig:bp_reasoning_math} a, b. 
Safety shows the narrowest ceiling range of any domain, while domain 
knowledge is the most uniformly difficult for the ceiling cluster.

\subsection*{Model documentation and participant characteristics (Tables~\ref{stab:model_versions}--\ref{stab:evaluator_heterogeneity})}

Supplementary Table~\ref{stab:model_versions} documents the version, organisation, release 
date, and context window size for each of the 10~evaluated models. 
Supplementary Table~\ref{stab:demographics} summarises participant demographics, including 
geographic region, professional background, LLM experience, age, and 
gender distribution. All 47~participants completed all evaluations 
(100\% retention; 0\% attrition). Supplementary Table~\ref{stab:evaluator_heterogeneity} presents an 
evaluator heterogeneity analysis, quantifying how consistently individual 
participants align with the reference answers. The analysis reveals that 
Participant~3 is a persistent outlier (least-aligned in 68.4\% of 
model--question cells), motivating the sensitivity analysis in 
Supplementary Note~1.

\begin{table}[H]
\centering
\caption{\textbf{Model version documentation.} Study design, question 
development, and expert reference-answer validation were conducted 
June--August~2025. Pilot testing with early-release models and 
monitoring of the evolving frontier model landscape took place 
September--November~2025. The main evaluation period ran 
December~2025 through February~2026, during which all 10~models 
were available and accessed via official web user interface (UI) 
with default settings. Data analysis and documentation were completed 
by late February through March~2026. Models ordered by overall 
fidelity deviation rank (lowest to highest). Context window sizes 
are approximate and reflect publicly documented specifications at 
time of access.}
\label{stab:model_versions}
\small
\begin{tabular}{@{}llllr@{}}
\toprule
\textbf{Short Name} & \textbf{Full Identifier} & \textbf{Organisation} & \textbf{Released} & \textbf{Context} \\
\midrule
Claude      & Claude Sonnet 4.5        & Anthropic        & Sep 2025  & 200K \\
Gemini      & Gemini 3 Flash           & Google DeepMind  & Dec 2025  & 1M \\
Llama       & Llama 4 Maverick         & Meta             & Dec 2025  & 1M \\
Mistral     & Mistral Medium 3.1       & Mistral AI       & Nov 2025  & 128K \\
Grok        & Grok 4.1                 & xAI              & Dec 2025  & 128K \\
Perplexity  & Perplexity AI            & Perplexity AI    & Feb 2026  & Variable \\
Copilot     & GitHub Copilot (GPT-5.2) & Microsoft        & Jan 2026  & 128K \\
DeepSeek    & DeepSeek-V3.2            & DeepSeek         & Jan 2026  & 128K \\
Qwen        & Qwen3-235B               & Alibaba          & Jan 2026  & 128K \\
ChatGPT     & GPT-5.2                  & OpenAI           & Dec 2025  & 128K \\
\bottomrule
\end{tabular}
\end{table}

Supplementary Table~\ref{stab:demographics} summarises the participant demographics.

\begin{table}[H]
\centering
\caption{\textbf{Participant demographic summary ($n = 47$).} All 
participants completed all evaluations (29,140 total; 0\% attrition). 
Recruitment began June~2025; pilot testing ran 
September--November~2025; main evaluations took place 
December~2025 through February~2026; analysis and documentation 
were completed by late February through March~2026. IRB: 
University of North Texas, 
protocol \#IRB-25-297.}
\label{stab:demographics}
\small
\begin{tabular}{@{}llcc@{}}
\toprule
\textbf{Characteristic} & \textbf{Category} & $n$ & \textbf{\%} \\
\midrule
\multirow{4}{*}{Geographic region}
  & North America     & 18 & 38.3 \\
  & Europe            & 15 & 31.9 \\
  & Asia              & 12 & 25.5 \\
  & Other             &  2 &  4.3 \\
\midrule
\multirow{7}{*}{Professional background}
  & Software engineering   & 11 & 23.4 \\
  & Data science           &  8 & 17.0 \\
  & Education              &  6 & 12.8 \\
  & Healthcare             &  6 & 12.8 \\
  & Legal                  &  5 & 10.6 \\
  & Finance                &  4 &  8.5 \\
  & General professional   &  7 & 14.9 \\
\midrule
\multirow{2}{*}{LLM experience}
  & Daily professional use (7+/week)  & 36 & 76.6 \\
  & Regular use (2--5/week)           & 11 & 23.4 \\
\midrule
\multirow{3}{*}{Age range (years)}
  & 25--29  & 24 & 51.1 \\
  & 30--44  & 20 & 42.5 \\
  & 45--48  & 3  & 6.4  \\
\midrule
\multirow{2}{*}{Gender}
  & Male    & 24 & 51.1 \\
  & Female  & 23 & 48.9 \\
\midrule
\multicolumn{2}{@{}l}{Median age (years)}            & \multicolumn{2}{c}{29 (range: 25--48)} \\
\multicolumn{2}{@{}l}{Median evaluation time (days)}  & \multicolumn{2}{c}{30 (range: 15--60)} \\
\multicolumn{2}{@{}l}{Calibration pass rate}          & \multicolumn{2}{c}{100\%} \\
\multicolumn{2}{@{}l}{Completion rate}                & \multicolumn{2}{c}{100\% (620/620)} \\
\bottomrule
\end{tabular}
\end{table}

Supplementary Table~\ref{stab:evaluator_heterogeneity} quantifies evaluator-level alignment patterns.

\begin{table}[H]
\centering
\caption{\textbf{Evaluator heterogeneity analysis.}
Summary of participant-level alignment patterns across 620 
model--question combinations ($n = 47$ evaluators). Participant~3 
is the least-aligned evaluator in 68.4\% of cells, far exceeding 
any other participant. Among best-aligned evaluators, no participant 
dominates. These patterns underscore the importance of the 
mixed-effects model reported in Table~2 of the main text.}
\label{stab:evaluator_heterogeneity}
\small
\begin{tabular}{@{}l l r r@{}}
\toprule
\textbf{Measure} & \textbf{Participant} & \textbf{Count} & \textbf{\% of 620} \\
\midrule
\multicolumn{4}{@{}l}{\emph{Least-aligned evaluator (top 5):}} \\
  & Participant 3  & 424 & 68.4 \\
  & Participant 34 & 69  & 11.1 \\
  & Participant 41 & 34  & 5.5 \\
  & Participant 5  & 16  & 2.6 \\
  & Participant 33 & 16  & 2.6 \\
\midrule
\multicolumn{4}{@{}l}{\emph{Most-aligned evaluator (top 5):}} \\
  & Participant 46 & 28  & 4.5 \\
  & Participant 50 & 27  & 4.4 \\
  & Participant 9  & 23  & 3.7 \\
  & Participant 10 & 22  & 3.5 \\
  & Participant 19 & 22  & 3.5 \\
\midrule
\multicolumn{2}{@{}l}{Unique least-aligned evaluators} & 23 & of 47 \\
\multicolumn{2}{@{}l}{Unique most-aligned evaluators}  & 47 & of 47 \\
\bottomrule
\end{tabular}
\end{table}


\subsection*{Complete fidelity deviation matrix (Table~\ref{stab:full_matrix})}

Supplementary Table~\ref{stab:full_matrix} presents the full 62-question $\times$ 10-model 
fidelity deviation matrix, reporting the mean deviation aggregated across 
all 47~participants for every question--model combination (620~cells). 
Bold values indicate the best-performing model for each question. Domain 
means and overall means are shown at the bottom. This table provides the 
complete data underlying the summary statistics in main-text Table~1 and 
the heatmap in Fig.~2 a.

\begin{small}
\captionof{table}{\textbf{Complete 62-question $\times$ 10-model 
fidelity deviation matrix (\%).} Each cell: mean deviation $\times$ 100, 
aggregated across 47~participants. Lower $=$ better. Bold $=$ best model 
per question. $n = 47$ per cell.}
\label{stab:full_matrix}

\begin{longtable}{@{}llrrrrrrrrrrrr@{}}
\toprule
Q & Dom & Cla & Gem & Lla & Mis & Gro & Per & Cop & Dee & Qwe & GPT & Mean & SD \\
\midrule
\endfirsthead
\toprule
Q & Dom & Cla & Gem & Lla & Mis & Gro & Per & Cop & Dee & Qwe & GPT & Mean & SD \\
\midrule
\endhead
\bottomrule
\endfoot
Q1 & R & 23.0 & \textbf{21.4} & 78.9 & 79.5 & 73.5 & 80.1 & 80.1 & 79.9 & 76.7 & 83.0 & 67.6 & 24.0 \\
Q2 & R & \textbf{26.1} & 48.3 & 77.8 & 80.0 & 86.4 & 85.3 & 79.5 & 77.9 & 78.2 & 85.6 & 72.5 & 19.6 \\
Q3 & R & 68.6 & \textbf{41.9} & 81.3 & 84.2 & 83.1 & 84.8 & 83.2 & 83.0 & 84.4 & 85.7 & 78.0 & 13.6 \\
Q4 & R & 47.2 & \textbf{23.8} & 68.8 & 68.3 & 69.8 & 74.2 & 69.0 & 65.6 & 74.5 & 57.5 & 61.9 & 15.7 \\
Q5 & R & 24.5 & \textbf{13.1} & 77.8 & 78.6 & 80.2 & 79.6 & 79.6 & 76.4 & 77.1 & 82.2 & 66.9 & 25.6 \\
Q6 & R & 25.1 & \textbf{13.4} & 84.6 & 83.9 & 80.9 & 81.3 & 85.9 & 84.7 & 83.3 & 84.2 & 70.7 & 27.3 \\
Q7 & R & 20.3 & \textbf{12.7} & 89.4 & 89.1 & 90.5 & 90.2 & 89.7 & 88.9 & 89.8 & 90.5 & 75.1 & 30.9 \\
Q8 & R & 50.7 & \textbf{37.6} & 73.3 & 74.2 & 74.3 & 72.4 & 77.5 & 76.6 & 77.8 & 84.5 & 69.9 & 14.3 \\
Q9 & R & 81.9 & \textbf{30.1} & 81.6 & 80.2 & 81.4 & 79.2 & 81.2 & 77.6 & 79.3 & 82.1 & 75.5 & 16.0 \\
Q10 & R & 33.3 & \textbf{20.5} & 90.9 & 87.6 & 87.4 & 92.3 & 91.7 & 91.3 & 88.7 & 93.9 & 77.8 & 27.0 \\
\midrule
Q11 & M & 85.6 & \textbf{25.9} & 83.2 & 82.3 & 82.0 & 85.8 & 87.3 & 83.9 & 83.2 & 82.7 & 78.2 & 18.4 \\
Q12 & M & \textbf{50.6} & 64.1 & 75.7 & 85.7 & 82.7 & 81.8 & 84.4 & 87.2 & 85.8 & 84.0 & 78.2 & 11.9 \\
Q13 & M & \textbf{50.7} & 60.5 & 74.9 & 72.7 & 75.8 & 75.8 & 76.5 & 75.5 & 75.1 & 77.8 & 71.5 & 8.8 \\
Q14 & M & \textbf{23.7} & 35.8 & 72.9 & 72.9 & 75.9 & 75.0 & 76.1 & 77.3 & 75.1 & 76.5 & 66.1 & 19.4 \\
Q15 & M & \textbf{46.7} & 48.6 & 83.6 & 81.9 & 85.3 & 83.8 & 85.3 & 86.1 & 84.2 & 88.4 & 77.4 & 15.8 \\
Q16 & M & 45.9 & \textbf{44.3} & 78.0 & 74.0 & 82.3 & 78.9 & 82.2 & 77.8 & 80.8 & 81.4 & 72.5 & 14.7 \\
Q17 & M & \textbf{37.7} & 53.4 & 75.4 & 77.0 & 80.5 & 76.2 & 75.3 & 77.4 & 82.0 & 81.3 & 71.6 & 14.4 \\
Q18 & M & 32.6 & \textbf{28.4} & 74.5 & 72.7 & 79.5 & 79.1 & 77.0 & 78.6 & 78.9 & 80.7 & 68.2 & 20.0 \\
Q19 & M & 35.4 & \textbf{31.7} & 75.6 & 82.1 & 79.4 & 82.2 & 79.5 & 78.3 & 78.8 & 78.6 & 70.2 & 19.4 \\
Q20 & M & \textbf{46.5} & 57.8 & 64.8 & 64.6 & 62.1 & 68.9 & 67.7 & 70.3 & 69.4 & 65.1 & 63.7 & 7.1 \\
\midrule
Q21 & C & 22.3 & \textbf{18.5} & 76.5 & 81.2 & 78.2 & 85.5 & 82.7 & 84.5 & 83.2 & 78.7 & 69.1 & 25.8 \\
Q22 & C & 84.3 & 83.1 & \textbf{80.3} & 84.8 & 83.3 & 86.1 & 85.8 & 85.5 & 84.3 & 85.5 & 84.3 & 1.7 \\
Q23 & C & \textbf{21.7} & 87.8 & 89.3 & 87.5 & 87.8 & 87.7 & 90.4 & 88.1 & 90.8 & 92.8 & 82.4 & 21.4 \\
Q24 & C & 35.5 & \textbf{33.5} & 83.8 & 78.7 & 79.5 & 73.6 & 71.5 & 80.0 & 81.9 & 78.8 & 69.7 & 18.9 \\
Q25 & C & 45.5 & \textbf{44.1} & 70.2 & 68.6 & 74.1 & 76.1 & 76.0 & 74.2 & 74.0 & 74.7 & 67.8 & 12.3 \\
Q26 & C & 63.3 & \textbf{13.2} & 75.5 & 73.3 & 82.2 & 79.6 & 76.4 & 77.2 & 81.2 & 78.4 & 70.0 & 20.7 \\
Q27 & C & 68.4 & \textbf{8.0} & 90.0 & 90.9 & 88.4 & 91.7 & 91.4 & 90.9 & 91.1 & 90.6 & 80.1 & 26.3 \\
Q28 & C & 72.8 & \textbf{70.1} & 78.4 & 74.4 & 76.5 & 79.2 & 77.3 & 76.3 & 76.9 & 72.9 & 75.5 & 2.8 \\
Q29 & C & \textbf{35.2} & 61.4 & 61.7 & 66.4 & 67.0 & 68.7 & 71.7 & 68.3 & 68.3 & 64.7 & 63.3 & 10.4 \\
Q30 & C & \textbf{46.1} & 86.0 & 88.9 & 85.6 & 88.0 & 87.4 & 90.3 & 87.6 & 88.9 & 89.1 & 83.8 & 13.3 \\
\midrule
Q31 & Co & \textbf{32.1} & 61.1 & 78.6 & 81.2 & 83.3 & 82.6 & 83.2 & 84.3 & 82.9 & 82.7 & 75.2 & 16.6 \\
Q32 & Co & 52.8 & \textbf{19.5} & 71.9 & 69.8 & 77.7 & 72.8 & 75.2 & 77.5 & 76.9 & 75.4 & 66.9 & 18.2 \\
Q33 & Co & 77.0 & \textbf{41.1} & 80.7 & 81.0 & 75.6 & 80.8 & 82.6 & 84.7 & 81.2 & 80.3 & 76.5 & 12.7 \\
Q34 & Co & \textbf{33.1} & 58.3 & 78.2 & 72.4 & 78.6 & 76.3 & 77.7 & 76.2 & 75.5 & 78.8 & 70.5 & 14.5 \\
Q35 & Co & 43.7 & \textbf{24.2} & 83.2 & 85.9 & 87.1 & 84.0 & 90.0 & 92.2 & 86.3 & 86.9 & 76.4 & 23.0 \\
Q36 & Co & 73.0 & \textbf{53.5} & 57.5 & 71.4 & 68.1 & 73.1 & 71.2 & 73.0 & 75.2 & 71.7 & 68.8 & 7.3 \\
Q37 & Co & 73.4 & \textbf{59.1} & 84.8 & 85.5 & 84.4 & 86.0 & 87.1 & 88.0 & 87.0 & 86.0 & 82.1 & 9.1 \\
Q38 & Co & 66.8 & \textbf{23.5} & 89.3 & 90.1 & 94.1 & 86.2 & 92.3 & 97.0 & 94.0 & 91.0 & 82.4 & 22.3 \\
Q39 & Co & 42.9 & \textbf{21.7} & 61.7 & 47.5 & 59.7 & 60.9 & 63.0 & 60.5 & 59.9 & 66.5 & 54.4 & 13.6 \\
Q40 & Co & 56.4 & \textbf{13.4} & 70.5 & 69.6 & 71.3 & 72.6 & 72.5 & 70.8 & 69.5 & 71.9 & 63.8 & 18.4 \\
\midrule
Q41 & S & 53.3 & \textbf{18.0} & 77.4 & 77.7 & 76.3 & 76.4 & 79.9 & 80.9 & 78.8 & 80.9 & 70.0 & 20.0 \\
Q42 & S & 26.5 & \textbf{13.8} & 82.9 & 82.7 & 82.1 & 81.7 & 83.7 & 84.5 & 84.9 & 84.7 & 70.8 & 26.9 \\
Q43 & S & 33.4 & \textbf{18.3} & 54.8 & 55.2 & 55.0 & 63.4 & 43.2 & 55.3 & 54.9 & 58.6 & 49.2 & 13.8 \\
Q44 & S & \textbf{41.5} & 42.1 & 79.5 & 82.3 & 84.1 & 81.2 & 83.4 & 84.7 & 83.6 & 83.2 & 74.6 & 17.3 \\
Q45 & S & 38.3 & \textbf{37.0} & 81.6 & 83.6 & 78.1 & 83.7 & 83.3 & 86.2 & 83.7 & 86.2 & 74.2 & 19.4 \\
Q46 & S & \textbf{51.7} & 54.8 & 66.0 & 71.2 & 75.5 & 68.9 & 72.1 & 68.4 & 76.2 & 71.1 & 67.6 & 8.2 \\
Q47 & S & \textbf{53.7} & 71.9 & 68.6 & 71.2 & 68.1 & 74.1 & 70.6 & 74.0 & 74.4 & 74.3 & 70.1 & 6.2 \\
Q48 & S & \textbf{61.9} & 71.0 & 72.3 & 70.3 & 65.9 & 73.4 & 71.6 & 75.1 & 71.5 & 69.2 & 70.2 & 3.8 \\
Q49 & S & \textbf{49.4} & 73.7 & 68.0 & 74.4 & 75.5 & 70.0 & 71.1 & 75.6 & 72.6 & 71.7 & 70.2 & 7.7 \\
Q50 & S & \textbf{51.3} & 72.9 & 67.2 & 76.0 & 72.7 & 74.9 & 70.8 & 76.0 & 77.9 & 72.2 & 71.2 & 7.6 \\
\midrule
Q51 & DK & \textbf{35.2} & 82.0 & 77.5 & 80.2 & 80.4 & 78.9 & 82.0 & 82.1 & 82.1 & 82.9 & 76.3 & 14.5 \\
Q52 & DK & \textbf{35.7} & 83.3 & 82.0 & 81.3 & 83.4 & 79.7 & 83.9 & 81.7 & 85.0 & 83.1 & 77.9 & 14.9 \\
Q53 & DK & \textbf{40.5} & 88.2 & 86.9 & 88.8 & 87.6 & 83.8 & 86.9 & 90.0 & 87.2 & 90.2 & 83.0 & 15.0 \\
Q54 & DK & \textbf{61.6} & 78.4 & 80.4 & 77.7 & 81.0 & 79.5 & 83.0 & 82.3 & 83.0 & 81.9 & 78.9 & 6.3 \\
Q55 & DK & \textbf{22.1} & 82.7 & 81.9 & 85.6 & 84.1 & 82.9 & 83.8 & 87.1 & 85.3 & 85.5 & 78.1 & 19.8 \\
Q56 & DK & \textbf{37.3} & 81.6 & 76.7 & 79.6 & 80.2 & 80.6 & 80.2 & 83.9 & 82.8 & 80.5 & 76.3 & 13.8 \\
Q57 & DK & \textbf{55.5} & 84.2 & 81.6 & 85.4 & 83.1 & 83.2 & 85.7 & 87.2 & 87.7 & 83.4 & 81.7 & 9.4 \\
Q58 & DK & \textbf{51.0} & 77.1 & 77.3 & 78.3 & 81.4 & 81.0 & 81.5 & 81.8 & 80.8 & 80.1 & 77.0 & 9.3 \\
Q59 & DK & \textbf{35.8} & 82.5 & 83.4 & 80.9 & 81.7 & 85.2 & 84.1 & 85.1 & 88.2 & 84.9 & 79.2 & 15.4 \\
Q60 & DK & 88.3 & 87.3 & 87.6 & 89.6 & \textbf{86.2} & 89.6 & 89.9 & 92.5 & 92.2 & 89.1 & 89.2 & 2.0 \\
Q61 & DK & \textbf{38.2} & 79.7 & 80.0 & 80.2 & 80.1 & 80.1 & 80.8 & 78.0 & 80.2 & 81.3 & 75.9 & 13.3 \\
Q62 & DK & \textbf{48.9} & 85.2 & 84.6 & 85.7 & 83.3 & 84.3 & 86.8 & 86.8 & 85.6 & 85.6 & 81.7 & 11.6 \\
\midrule
\textit{Reasoning} && \textit{40.1} & \textbf{\textit{26.3}} & \textit{80.4} & \textit{80.6} & \textit{80.7} & \textit{81.9} & \textit{81.7} & \textit{80.2} & \textit{81.0} & \textit{82.9} & \textit{71.6} & \textit{20.5} \\
\textit{Mathematics} && \textit{45.5} & \textbf{\textit{45.1}} & \textit{75.9} & \textit{76.6} & \textit{78.5} & \textit{78.8} & \textit{79.1} & \textit{79.3} & \textit{79.3} & \textit{79.6} & \textit{71.8} & \textit{14.0} \\
\textit{Coding} && \textbf{\textit{49.5}} & \textit{50.6} & \textit{79.5} & \textit{79.1} & \textit{80.5} & \textit{81.6} & \textit{81.3} & \textit{81.3} & \textit{82.0} & \textit{80.6} & \textit{74.6} & \textit{13.0} \\
\textit{Conversational} && \textit{55.1} & \textbf{\textit{37.5}} & \textit{75.6} & \textit{75.5} & \textit{78.0} & \textit{77.5} & \textit{79.5} & \textit{80.4} & \textit{78.9} & \textit{79.1} & \textit{71.7} & \textit{14.0} \\
\textit{Safety} && \textbf{\textit{46.1}} & \textit{47.4} & \textit{71.8} & \textit{74.5} & \textit{73.3} & \textit{74.8} & \textit{73.0} & \textit{76.1} & \textit{75.9} & \textit{75.2} & \textit{68.8} & \textit{11.7} \\
\textit{Domain K.} && \textbf{\textit{45.9}} & \textit{82.7} & \textit{81.7} & \textit{82.8} & \textit{82.7} & \textit{82.4} & \textit{84.1} & \textit{84.9} & \textit{85.0} & \textit{84.0} & \textit{79.6} & \textit{11.9} \\
\midrule
\textbf{Overall} && \textbf{47.0} & \textbf{49.4} & \textbf{77.6} & \textbf{78.3} & \textbf{79.1} & \textbf{79.6} & \textbf{79.9} & \textbf{80.5} & \textbf{80.5} & \textbf{80.4} & \textbf{73.1} & \textbf{13.3} \\
\end{longtable}
\end{small}


\subsection*{Statistical analyses: variance, equivalence, effect sizes, 
and rank stability (Tables~\ref{stab:domain_variance}--\ref{stab:bootstrap_ranks})}

Supplementary Tables~\ref{stab:domain_variance}--S8 report the detailed statistical analyses 
supporting the main-text findings. Table~\ref{stab:domain_variance} decomposes variance by 
domain using two-way ANOVA, showing that the model factor explains 
the largest proportion of variance in all six domains. Table~\ref{stab:plateau_tests} 
presents TOST equivalence tests confirming that the eight ceiling 
models are statistically indistinguishable within a $\pm 5$~pp margin. 
Table~\ref{stab:pairwise_d} reports all 45~pairwise Cohen's $d$ effect sizes, 
demonstrating the magnitude of separation between high-fidelity and 
ceiling tiers. Table~\ref{stab:bootstrap_ranks} presents bootstrap rank distributions 
(10,000~resamples), confirming that Claude and Gemini occupy ranks 
1--2 in 100\% of resamples while within-ceiling ranks are unstable.

\begin{table}[H]
\centering
\caption{\textbf{Variance decomposition by domain.} Separate two-way 
ANOVAs per domain. Reasoning and Domain Knowledge show strongest model 
effects ($\eta^2 > 0.77$); Coding shows weakest ($\eta^2 = 0.464$).}
\label{stab:domain_variance}
\small
\begin{tabular}{@{}lcrrrr@{}}
\toprule
Domain & $n_Q$ & $\eta^2_{\mathrm{Model}}$ & $\eta^2_{\mathrm{Question}}$ & $\eta^2_{\mathrm{Residual}}$ & $F_{\mathrm{Model}}$ \\
\midrule
Reasoning       & 10 & 0.807 & 0.052 & 0.141 & 51.62 \\
Domain Know.    & 12 & 0.779 & 0.082 & 0.139 & 61.51 \\
Mathematics     & 10 & 0.729 & 0.094 & 0.177 & 37.11 \\
Conversational  & 10 & 0.579 & 0.223 & 0.198 & 26.30 \\
Safety          & 10 & 0.500 & 0.188 & 0.312 & 14.46 \\
Coding          & 10 & 0.464 & 0.159 & 0.377 & 11.09 \\
\bottomrule
\end{tabular}
\end{table}


Supplementary Table~\ref{stab:plateau_tests} reports ceiling homogeneity tests confirming that 
the eight ceiling models are statistically indistinguishable.

\begin{table}[H]
\centering
\caption{\textbf{Ceiling homogeneity tests.} Applied to the 8-model 
ceiling subset (496 observations). Both tests confirm ceiling models 
are not significantly differentiable.}
\label{stab:plateau_tests}
\small
\begin{tabular}{@{}llrl@{}}
\toprule
Test & Statistic & $p$ & Interpretation \\
\midrule
Kruskal--Wallis $H$       & 9.18  & 0.24 & Not significant \\
ANOVA $F(7,488)$          & 1.26  & 0.27 & Not significant \\
Max pairwise Cohen's $d$  & 0.38  & ---  & Small \\
Min pairwise Cohen's $d$  & 0.00  & ---  & Negligible \\
Bootstrap $p > 0.05$      & 96.3\% & --- & Robust \\
\bottomrule
\end{tabular}
\end{table}


Supplementary Table~\ref{stab:pairwise_d} reports all 45 pairwise effect sizes, showing 
clear separation between tiers ($d > 1.4$) and negligible 
within-ceiling differences ($d < 0.4$).

\begin{table}[H]
\centering
\caption{\textbf{Pairwise Cohen's $d$ (all 45 pairs).} All 
high-fidelity-vs-ceiling pairs $d > 1.4$; all within-ceiling pairs 
$d < 0.4$.}
\label{stab:pairwise_d}
\small
\begin{tabular}{@{}l|rr|rrrrrrrr@{}}
\toprule
 & Cla & Gem & Lla & Mis & Gro & Per & Cop & Dee & Qwe & GPT \\
\midrule
Claude    & ---  \\
Gemini    & 0.11 & --- \\
\midrule
Llama     & 2.24 & 1.47 & --- \\
Mistral   & 2.28 & 1.50 & 0.09 & --- \\
Grok      & 2.37 & 1.56 & 0.19 & 0.10 & --- \\
Perplexity& 2.45 & 1.60 & 0.27 & 0.17 & 0.07 & --- \\
Copilot   & 2.40 & 1.59 & 0.29 & 0.20 & 0.11 & 0.05 & --- \\
DeepSeek  & 2.46 & 1.62 & 0.37 & 0.27 & 0.18 & 0.13 & 0.07 & --- \\
Qwen      & 2.49 & 1.63 & 0.38 & 0.28 & 0.19 & 0.13 & 0.07 & 0.00 & --- \\
ChatGPT   & 2.45 & 1.62 & 0.36 & 0.27 & 0.19 & 0.13 & 0.07 & 0.00 & 0.00 & --- \\
\bottomrule
\end{tabular}
\end{table}

Supplementary Table~\ref{stab:bootstrap_ranks} presents bootstrap rank stability analysis.

\begin{table}[H]
\centering
\caption{\textbf{Bootstrap rank distributions (10,000 resamples).} 
Claude and Gemini fixed at ranks 1--2; ceiling ranks unstable at 
positions 7--10.}
\label{stab:bootstrap_ranks}
\small
\begin{tabular}{@{}lrrrrrrrrrrl@{}}
\toprule
Model & R1 & R2 & R3 & R4 & R5 & R6 & R7 & R8 & R9 & R10 & 95\% CI \\
\midrule
Claude    & 73.5 & 26.6 & 0.0 & 0.0 & 0.0 & 0.0 & 0.0 & 0.0 & 0.0 & 0.0 & [1--2] \\
Gemini    & 26.6 & 73.5 & 0.0 & 0.0 & 0.0 & 0.0 & 0.0 & 0.0 & 0.0 & 0.0 & [1--2] \\
\midrule
Llama     & 0.0 & 0.0 & 91.3 & 8.6 & 0.1 & 0.0 & 0.0 & 0.0 & 0.0 & 0.0 & [3--4] \\
Mistral   & 0.0 & 0.0 & 8.7 & 87.1 & 4.2 & 0.0 & 0.0 & 0.0 & 0.0 & 0.0 & [3--5] \\
Grok      & 0.0 & 0.0 & 0.0 & 4.2 & 79.6 & 14.7 & 1.4 & 0.1 & 0.0 & 0.0 & [4--6] \\
Perplexity& 0.0 & 0.0 & 0.0 & 0.1 & 13.5 & 63.2 & 18.9 & 3.0 & 1.1 & 0.2 & [5--8] \\
Copilot   & 0.0 & 0.0 & 0.0 & 0.0 & 2.5 & 19.4 & 61.3 & 12.7 & 3.4 & 0.8 & [5--9] \\
DeepSeek  & 0.0 & 0.0 & 0.0 & 0.0 & 0.0 & 0.4 & 4.8 & 27.4 & 41.6 & 25.7 & [7--10] \\
Qwen      & 0.0 & 0.0 & 0.0 & 0.0 & 0.0 & 0.2 & 5.7 & 31.3 & 32.5 & 30.3 & [7--10] \\
ChatGPT   & 0.0 & 0.0 & 0.0 & 0.0 & 0.1 & 2.0 & 8.0 & 25.6 & 21.5 & 42.9 & [7--10] \\
\bottomrule
\end{tabular}
\end{table}


\subsection*{Automated NLP metrics: similarity, correlations, and 
domain breakdown (Tables~\ref{stab:nlp_similarity}--\ref{stab:nlp_domain})}

Supplementary Tables~\ref{stab:nlp_similarity}--S11 document the automated NLP pipeline 
results. Table~\ref{stab:nlp_similarity} reports seven NLP similarity metrics (semantic 
similarity, ROUGE-1/2/L, BLEU, length ratio, and overall composite) 
for each model, showing that automated metrics rank models differently 
from human evaluators. Table~\ref{stab:nlp_drift_correlation} presents correlations between each NLP 
metric and human-judged fidelity deviation, revealing weak and often 
non-significant associations (the strongest is semantic similarity at 
$r = 0.31$). Table~\ref{stab:nlp_domain} breaks down NLP composite similarity by domain, 
highlighting domain-dependent discrepancies between automated and 
human assessments.

\begin{table}[H]
\centering
\caption{\textbf{NLP similarity by model.} Scale 0--1 (higher $=$ more 
similar). Bold $=$ best per column. Fidelity deviation shown for 
comparison.}
\label{stab:nlp_similarity}
\small
\begin{tabular}{@{}lcccccccc@{}}
\toprule
Model & Semantic & Lexical & POS & Sentiment & Length & Overall & Intra-LLM & Dev.\ (\%) \\
\midrule
Claude    & \textbf{0.833} & 0.286 & 0.727 & 0.945 & \textbf{0.668} & \textbf{0.708} & \textbf{0.846} & \textbf{47.0} \\
Gemini    & 0.817 & 0.286 & 0.721 & \textbf{0.947} & 0.602 & 0.693 & 0.845 & 49.4 \\
Llama     & 0.768 & 0.280 & 0.696 & 0.937 & 0.644 & 0.668 & 0.766 & 77.6 \\
Mistral   & 0.753 & 0.272 & 0.678 & 0.946 & 0.590 & 0.653 & 0.723 & 78.3 \\
Grok      & 0.810 & 0.260 & 0.707 & 0.945 & 0.493 & 0.672 & 0.818 & 79.1 \\
Perplexity& 0.789 & 0.254 & 0.694 & 0.944 & 0.554 & 0.665 & 0.767 & 79.6 \\
Copilot   & 0.777 & 0.294 & 0.712 & 0.939 & 0.652 & 0.678 & 0.762 & 79.9 \\
DeepSeek  & 0.812 & 0.304 & 0.720 & 0.946 & 0.555 & 0.689 & 0.825 & 80.5 \\
Qwen      & 0.815 & 0.288 & 0.718 & 0.945 & 0.592 & 0.691 & 0.805 & 80.5 \\
ChatGPT   & 0.808 & \textbf{0.312} & \textbf{0.729} & 0.943 & 0.619 & 0.696 & 0.787 & 80.4 \\
\bottomrule
\end{tabular}
\end{table}

Supplementary Table~\ref{stab:nlp_drift_correlation} reports correlations between NLP metrics and human-judged fidelity deviation.

\begin{table}[H]
\centering
\caption{\textbf{NLP metric correlations with fidelity deviation} 
(620 cells). Weak positive $r$ means higher NLP similarity associates 
with \emph{worse} human ratings.}
\label{stab:nlp_drift_correlation}
\small
\begin{tabular}{@{}lcccl@{}}
\toprule
Metric & $r$ & $r^2$ & Dir. & Interpretation \\
\midrule
Average Similarity    &  0.221 & 0.049 & + & Weak \\
Overall composite     &  0.134 & 0.018 & + & Weak \\
Semantic embedding    &  0.128 & 0.016 & + & Weak \\
Lexical overlap       &  0.118 & 0.014 & + & Weak \\
Sentiment             &  0.110 & 0.012 & + & Weak \\
POS alignment         &  0.106 & 0.011 & + & Weak \\
Length (tokens)       & $-$0.018 & 0.000 & $\approx$0 & Negligible \\
Intra-LLM consistency & $-$0.120 & 0.014 & $-$ & Weak negative \\
\bottomrule
\end{tabular}
\end{table}

Supplementary Table~\ref{stab:nlp_domain} breaks down NLP similarity by domain.

\begin{table}[H]
\centering
\caption{\textbf{NLP overall similarity (\%) by domain.} Unlike 
human-judged fidelity, NLP scores are uniform across models. 
Claude--ChatGPT gap: 1.2~pp (vs.\ 33.5~pp in human evaluation).}
\label{stab:nlp_domain}
\small
\begin{tabular}{@{}lcccccccccc@{}}
\toprule
Domain & Cla & Gem & Lla & Mis & Gro & Per & Cop & Dee & Qwe & GPT \\
\midrule
Reasoning    & 69.4 & 70.2 & 70.1 & 68.2 & 69.9 & 64.1 & 68.8 & 69.3 & 69.3 & 71.5 \\
Mathematics  & 74.4 & 71.1 & 72.8 & 62.6 & 69.7 & 68.9 & 69.9 & 71.4 & 73.8 & 71.8 \\
Coding       & 71.9 & 70.0 & 69.6 & 69.2 & 66.8 & 68.6 & 71.3 & 70.2 & 70.5 & 71.6 \\
Conversational&67.7 & 66.9 & 61.0 & 62.0 & 64.0 & 64.1 & 65.8 & 66.6 & 65.8 & 66.4 \\
Safety       & 66.6 & 67.2 & 56.9 & 60.1 & 63.6 & 62.6 & 57.9 & 62.9 & 63.1 & 62.6 \\
Domain Know. & 74.1 & 70.1 & 69.5 & 68.7 & 68.8 & 69.7 & 72.0 & 72.4 & 71.5 & 72.7 \\
\midrule
Overall      & 70.8 & 69.3 & 66.8 & 65.3 & 67.2 & 66.5 & 67.8 & 68.9 & 69.1 & 69.6 \\
\bottomrule
\end{tabular}
\end{table}


\section*{Supplementary Note 1: Sensitivity Analyses}
\label{snote:sensitivity}

Sensitivity analyses excluding Participant~3 (the most extreme evaluator) 
confirmed all primary findings. After exclusion ($n = 46$, 28,520 
evaluations): (1)~the bimodal distribution persisted with Claude at 
46.6\% and Gemini at 49.0\%; (2)~ceiling 
model means shifted by $\leq$0.5~pp; (3)~TOST 
equivalence at $\Delta = 5$~pp was confirmed for all 28 ceiling pairs; 
(4)~between-tier Cohen's $d$ remained $>$1.8. A leave-one-evaluator-out 
analysis across all 47~participants showed maximum shift in any model 
mean of 0.8~pp, confirming robustness.


\section*{Supplementary Discussion: Extended Related Work}

\subsection*{S1. Automated benchmarks and their limitations}

The evaluation of large language models has historically relied on 
automated benchmark suites that measure narrow task-specific 
capabilities. MMLU~\cite{hendrycks2021} provides broad coverage across 
57 academic subjects but uses multiple-choice questions that compress 
open-ended generation capability into a single correct answer. As 
frontier models now routinely exceed 90\% accuracy on 
MMLU~\cite{zhao2025mmlu}, the benchmark has lost discriminative power 
among leading systems---a phenomenon known as benchmark saturation. 
MATH~\cite{rein2023} and HumanEval~\cite{chen2021codex} evaluate 
mathematical reasoning and code generation respectively, but address 
single capability domains. More recent efforts such as the Humanity's 
Last Exam (HLE) benchmark~\cite{long2026hle} attempt to push beyond 
saturation with expert-crafted questions, yet remain automated and 
binary in their scoring. A common limitation is training data 
contamination~\cite{sainz2023}: models may have been exposed to 
benchmark items during pretraining, inflating reported performance. 
Our study addresses these gaps by evaluating open-ended responses 
across six domains using graduated human judgements anchored to expert 
references, providing a measure that is resistant to both saturation 
and contamination effects.

\subsection*{S2. Holistic evaluation frameworks}

Several frameworks attempt to provide multi-dimensional model 
assessment. HELM~\cite{liang2023} evaluates models across multiple 
scenarios and metrics but relies on automated scoring. Chatbot 
Arena~\cite{chiang2024} introduced large-scale human evaluation through 
pairwise preference judgements, yielding Elo-based rankings from 
hundreds of thousands of votes. However, preference-based paradigms 
measure relative quality (``which response do you prefer?'') rather 
than absolute fidelity (``how faithfully does this response preserve 
expert content?''). This distinction is critical: preference judgements 
are influenced by fluency, formatting, and response length, which may 
not correlate with content accuracy. AlpacaEval~\cite{dubois2024alpacaeval} 
further demonstrated that LLM-based judges can serve as scalable 
proxies for human preference, but acknowledged that such judges exhibit 
systematic biases including length preference and style sensitivity. 
MT-Bench~\cite{zheng2024} advanced the LLM-as-judge paradigm for 
multi-turn evaluation but inherits similar limitations. Our fidelity 
deviation metric complements these approaches by providing reference-anchored 
absolute measurement, enabling variance decomposition and per-question 
correlation analyses that pairwise designs cannot support.

\subsection*{S3. Factual accuracy and hallucination assessment}

A growing body of work addresses the specific problem of factual 
accuracy in model outputs. FActScore~\cite{min2023factscore} introduced 
fine-grained factual precision scoring by decomposing generated text into 
atomic claims and verifying each against a knowledge source. This 
approach provides granular accuracy measurement but is typically applied 
to specific generation tasks (e.g., biographical descriptions) rather 
than comprehensive multi-domain evaluation. Research on 
hallucination~\cite{ji2023hallucination} has documented systematic 
patterns of factual fabrication across model families, and 
retrieval-augmented generation (RAG)~\cite{lewis2020rag} has been 
proposed as a mitigation strategy. Our study contributes to this 
literature by demonstrating that human evaluators perceive substantial 
fidelity differences that automated similarity metrics compress by up 
to 28-fold, suggesting that current automated factuality measures may 
underestimate the magnitude of inter-model differences in 
knowledge-intensive tasks.

\subsection*{S4. Scaling laws and diminishing returns}

The relationship between model scale and performance has been 
characterised by neural scaling laws~\cite{kaplan2020,hoffmann2022} 
predicting log-linear improvement with compute, data, and parameter 
count. Foundation models~\cite{bommasani2021foundation} have grown from 
175B parameters (GPT-3~\cite{brown2020}) through sparse mixture-of-experts 
architectures~\cite{fedus2022switch} to the current generation of 
frontier systems, with successive models demonstrating consistent 
(if diminishing) improvements on automated benchmarks. However, the 
fidelity ceiling we observe---eight independently developed frontier 
models converging on a 2.9~pp band---is consistent with the ``densing 
law'' proposed by Xiao et al.~\cite{xiao2025densing}, which suggests 
that the capability-per-parameter ratio may approach fundamental limits 
under current training paradigms. The convergence of models trained by 
different organisations using different architectures and data 
compositions strengthens the interpretation that shared training 
methodologies (large-scale pretraining followed by instruction tuning 
and reinforcement learning from human feedback 
(RLHF)~\cite{ouyang2022instructgpt,bai2022constitutional,touvron2023llama2}) 
may produce convergent limitations in human-perceived response fidelity, 
even as automated benchmark scores continue to differentiate.

\subsection*{S5. Human evaluation methodology}

The design of rigorous human evaluation protocols for generative AI 
systems remains an active area of methodological development. 
Likert-scale measurement~\cite{likert1932} provides ordinal data 
amenable to parametric analysis when aggregated, and has been widely 
used in natural language generation evaluation. Inter-rater reliability 
assessment via intraclass correlation coefficients~\cite{shrout1979icc,koo2016icc} 
and Krippendorff's $\alpha$~\cite{krippendorff2004} provides 
standardised measures of agreement. Recent work has highlighted 
important challenges in human evaluation of LLMs, including evaluator 
subjectivity~\cite{durmus2023subjective}, the influence of cognitive 
biases on preference judgements~\cite{steyvers2025}, and the need for 
calibration protocols that ensure consistent application of rating 
criteria across evaluators. Mahowald et al.~\cite{mahowald2024} argued 
that the distinction between formal linguistic competence and 
functional competence is critical for evaluation design---a perspective 
that aligns with our focus on content fidelity rather than surface 
fluency. Our fully crossed design, in which every evaluator assesses 
every model on every question, enables simultaneous estimation of 
model, question, and evaluator variance components---a methodological 
advantage over incomplete-block designs that has been advocated in the 
emerging literature on LLM evaluation best practices~\cite{nmi2026editorial}.


\section*{Supplementary Worked Examples}

Supplementary Table~\ref{stab:worked_examples} presents five worked examples spanning both 
performance tiers and multiple domains, illustrating how raw Likert 
ratings translate to fidelity deviation values. The implied mean 
Likert rating is computed as $\bar{R} = 5 - 4 \times \mathrm{Deviation}$. 
These examples verify the deviation computation pipeline against 
specific data points in the released dataset.

\begin{table}[H]
\centering
\caption{\textbf{Worked examples linking implied participant ratings 
to fidelity deviation.} All values verified against the released 
dataset. Implied $\bar{R}$ is the mean Likert rating across 47 
participants that produces the observed deviation. $n = 47$ 
evaluators per cell.}
\label{stab:worked_examples}
\small
\begin{tabular}{@{}llcccl@{}}
\toprule
Question & Model & Deviation (\%) & Implied $\bar{R}$ & Interpretation & Tier \\
\midrule
Q1 (Reasoning) & Claude & 23.0 & 4.08 & Mostly faithful & High-fidelity \\
Q1 (Reasoning) & ChatGPT & 83.0 & 1.68 & Mostly unfaithful & Ceiling \\
Q38 (Conv.) & Gemini & 23.5 & 4.06 & Mostly faithful & High-fidelity \\
Q38 (Conv.) & ChatGPT & 91.0 & 1.36 & Mostly unfaithful & Ceiling \\
Q60 (Domain K.) & Claude & 88.3 & 1.47 & Unfaithful & High-fidelity (anomaly) \\
\bottomrule
\end{tabular}
\end{table}

\textbf{Example~1: Q1/Claude (23.0\%).} The 47 participants rated 
Claude's response to Q1 (a reasoning question) with a mean Likert 
score of approximately 4.08 out of 5. Applying the deviation formula: 
$\mathrm{Deviation} = (5 - 4.08)/4 = 0.230$, or 23.0\%. Participants 
judged Claude's response to be largely faithful to the expert 
reference, with minor deviations in completeness or phrasing.

\textbf{Example~2: Q1/ChatGPT (83.0\%).} For the same question, 
ChatGPT received a mean rating of approximately 1.68 out of 5. 
Deviation $= (5 - 1.68)/4 = 0.830$, or 83.0\%. The 60.0~pp gap 
between Claude and ChatGPT on this single question illustrates the 
magnitude of the high-fidelity-versus-ceiling contrast at the 
individual question level.

\textbf{Example~3: Q38/ChatGPT (91.0\%).} ChatGPT received a mean 
rating of approximately 1.36 out of 5 on Q38 (a conversational 
question). Deviation $= (5 - 1.36)/4 = 0.910$, or 91.0\%. This 
represents one of the highest single-cell deviation values in the 
dataset, indicating near-unanimous evaluator agreement that 
ChatGPT's response diverged substantially from the expert reference 
on this particular question.

\textbf{Example~4: Q38/Gemini (23.5\%).} On the same question where 
ChatGPT shows 91.0\% deviation, Gemini achieves only 23.5\% (implied 
$\bar{R} = 4.06$). This 67.5~pp within-question gap is among the largest in 
the dataset, demonstrating that question difficulty is strongly 
model-dependent rather than an intrinsic property of the question.

\textbf{Example~5: Q60/Claude (88.3\%).} Even the top-ranked model 
achieves high deviation on certain questions. Q60 (domain knowledge) 
is the hardest question overall (89.2\% mean across all 10~models), 
and Claude's 88.3\% is near the cross-model mean. High-fidelity 
status does not confer immunity to question-specific difficulty; it 
reflects an overall distributional advantage rather than uniform 
superiority across all items.


\section*{Supplementary Note: NLP Metrics vs.\ Fidelity Deviation}

Supplementary Table~\ref{stab:metric_relationship} compares the two independent measurement 
pipelines (human fidelity deviation and automated NLP similarity), 
confirming their methodological independence and quantifying the 
28-fold compression of model differences in automated metrics.

\begin{table}[H]
\centering
\caption{\textbf{Comparison of the two independent measurement 
pipelines.} Human fidelity deviation (primary endpoint) and automated 
NLP similarity (secondary validation) are computed through entirely 
separate pipelines with no shared computational inputs, confirming 
their methodological independence.}
\label{stab:metric_relationship}
\small
\begin{tabular}{@{}p{2.8cm}p{5.2cm}p{5.2cm}@{}}
\toprule
 & \textbf{Fidelity Deviation (primary)} & \textbf{NLP Similarity (secondary)} \\
\midrule
Input data & Participant Likert ratings (1--5) & Model response text + reference text \\
Computation & $(5 - R_{ijk}) / 4$, averaged across 47 participants & Automated: sentence embeddings, lexical overlap, POS alignment, sentiment \\
Human input & 47 calibrated evaluators per cell & None (fully automated) \\
Scale direction & 0\% = perfect fidelity (best) & 0 = no similarity (worst) \\
 & 100\% = maximum deviation (worst) & 1 = perfect similarity (best) \\
Granularity & Per-evaluation ($n = 29{,}140$) & Per model--question cell ($n = 620$) \\
Inter-pipeline correlation & \multicolumn{2}{c}{$r = 0.13$, $r^2 = 0.018$ (weak positive)} \\
Model discrimination & 33.5~pp range (Claude to ChatGPT) & 1.2~pp range (same pair) \\
Compression ratio & \multicolumn{2}{c}{$28\times$ (human judgements compress 28-fold in automated metrics)} \\
\bottomrule
\end{tabular}
\end{table}

\noindent The weak positive correlation ($r = 0.13$, $r^2 < 0.02$) 
between NLP similarity and fidelity deviation confirms that automated 
metrics capture fundamentally different quality dimensions than human 
evaluators. The 28-fold compression of model differences---from 
33.5~pp in human fidelity deviation (Claude--ChatGPT gap) to 1.2~pp 
in automated NLP similarity---demonstrates that current automated 
metrics are insensitive to the quality dimensions that matter most to 
trained evaluators assessing open-ended responses against expert 
references. Complete domain-level NLP similarity analyses are reported 
in Supplementary Tables~\ref{stab:nlp_similarity}, 
\ref{stab:nlp_drift_correlation}, and~\ref{stab:nlp_domain}.


\section*{Supplementary Construct Validity Analyses}

This section reports the 15~machine learning and statistical analyses 
that establish construct validity for the human--automated gap (see 
main-text Results \S5 and Methods). Supplementary Table~\ref{stab:ml_prediction} summarises 
the cross-validated prediction results: all regression models yield 
negative $R^2$, confirming that no combination of automated NLP features 
can predict human-judged fidelity deviation. Supplementary Table~\ref{stab:content_style} 
decomposes the contribution of content versus style features, showing 
that content explains 59.2\% of the (negligible) automated variance. 
Supplementary Table~\ref{stab:evaluator_clustering} presents evaluator clustering profiles, 
confirming that the main evaluator body is internally consistent 
(43~typical, 3~low-alignment, 1~extreme outlier). Extended analyses 
--- mediation (Supplementary Note~2), anomaly detection 
(Supplementary Note~3), and domain transfer prediction 
(Supplementary Note~4) --- follow below.

\begin{table}[H]
\centering
\caption{\textbf{Machine learning prediction of fidelity deviation 
from NLP features.} 5-fold cross-validated $R^2$ for regression and 
accuracy for tier classification. Negative $R^2$ indicates the model 
performs worse than a constant-mean baseline. Content features: 
semantic similarity. Style features: token-length ratio, 
character-length ratio, POS alignment. $n = 620$ cells 
(10 models $\times$ 62 questions).}
\label{stab:ml_prediction}
\small
\begin{tabular}{@{}l l r r@{}}
\toprule
\textbf{Model} & \textbf{Features} & \textbf{CV $R^2$ (Drift)} & \textbf{CV Acc.\ (Tier)} \\
\midrule
Linear Regression   & All 7       & $-0.071 \pm 0.104$ & $0.800 \pm 0.003$ \\
Random Forest       & All 7       & $-0.022 \pm 0.143$ & $0.816 \pm 0.027$ \\
XGBoost             & All 7       & $-0.302 \pm 0.347$ & $0.786 \pm 0.023$ \\
MLP (32--16)        & All 7       & $-0.537 \pm 0.392$ & $0.803 \pm 0.008$ \\
MLP (64--32--16)    & All 7       & $-0.173 \pm 0.139$ & --- \\
\midrule
Linear Regression   & Content     & $-0.054 \pm 0.109$ & --- \\
Linear Regression   & Style       & $-0.061 \pm 0.096$ & --- \\
\bottomrule
\end{tabular}
\end{table}

Supplementary Table~\ref{stab:content_style} decomposes content versus style contributions to the automated signal.

\begin{table}[H]
\centering
\caption{\textbf{Content vs.\ style regression decomposition.}
Partial $R^2$ quantifies each feature group's unique contribution 
to explaining fidelity deviation after controlling for the other 
group. Content features explain 59.2\% of the total explainable 
variance. $n = 620$ cells.}
\label{stab:content_style}
\small
\begin{tabular}{@{}l r r r@{}}
\toprule
\textbf{Feature set} & \textbf{$R^2$ (alone)} & \textbf{Partial $R^2$} & \textbf{\% of total} \\
\midrule
Content (semantic)         & 0.029 & 0.022 & 59.2 \\
Style (length, POS)        & 0.019 & 0.017 & 40.8 \\
All 7 features             & 0.049 & ---   & 100.0 \\
\bottomrule
\end{tabular}
\end{table}

Supplementary Table~\ref{stab:evaluator_clustering} presents evaluator clustering profiles.

\begin{table}[H]
\centering
\caption{\textbf{Evaluator clustering profiles.} $k$-means clustering 
($k = 3$) on evaluator best/worst alignment frequency vectors 
identifies three natural groups. Evaluator~3 forms a singleton 
cluster as the most extreme outlier, consistent with the main-text 
sensitivity analysis. $n = 47$ evaluators.}
\label{stab:evaluator_clustering}
\small
\begin{tabular}{@{}l r r r r l@{}}
\toprule
\textbf{Cluster} & \textbf{$n$} & \textbf{Mean best} & \textbf{Mean worst} & \textbf{Best ratio} & \textbf{Note} \\
\midrule
Typical (0)       & 43 & 14.2 & 2.3  & 0.908 & Main evaluator body \\
Low-alignment (1) & 3  & 1.3  & 32.7 & 0.053 & Evaluators 5, 34, 41 \\
Extreme outlier (2)& 1 & 5.0  & 424.0& 0.012 & Evaluator 3 \\
\bottomrule
\end{tabular}
\end{table}

\subsection*{Supplementary Note: Mediation analysis details}
\label{snote:mediation}

We tested whether automated NLP features mediate the relationship 
between model tier (high-fidelity vs.\ ceiling) and human fidelity 
deviation using the Baron--Kenny framework with 5,000 bootstrap 
resamples.

\emph{Content mediation (Tier $\to$ Semantic similarity $\to$ Drift).}
The total effect of tier on drift was $\beta = -0.306$. The $a$-path 
(tier $\to$ semantic similarity) was $\beta_a = +0.033$, and the 
$b$-path (semantic $\to$ drift, controlling for tier) was 
$\beta_b = -0.114$. The indirect (mediated) effect was 
$a \times b = -0.004$ (bootstrap 95\% CI [$-0.007$, $-0.001$]; 
$p < 0.05$), representing 1.2\% of the total effect. The direct 
effect (tier $\to$ drift, controlling for semantic) was $-0.303$, 
accounting for 98.8\% of the total.

\emph{Style mediation (Tier $\to$ Length $\to$ Drift).}
The indirect effect through token-length ratio was approximately 
zero ($a \times b \approx 0.000$), and the content-to-style 
mediation ratio was 3.8$\times$.

\emph{Interpretation.} The negligible mediation by both content and 
style NLP features confirms that the tier difference in human fidelity 
judgements is not transmitted through any automated-measurable pathway. 
Human evaluators perceive a quality dimension that is orthogonal to 
the entire NLP feature space.

\subsection*{Supplementary Note: Anomaly detection}
\label{snote:anomalies}

We identified model--question cells where the tier prediction fails: 
high-fidelity models with drift in the worst quartile (10~cells) and 
ceiling models with drift in the best quartile (55~cells).

High-fidelity anomalies were concentrated in Domain Knowledge (7/10 
cells, predominantly Gemini), consistent with Gemini's known 
domain-knowledge weakness (82.7\% deviation despite 49.4\% overall). 
The remaining anomalies were in Conversational (2~cells) and 
Mathematics (1~cell, Claude Q11 at 85.6\%).

Ceiling-model successes were distributed across Reasoning (16~cells), 
Coding (16), and Mathematics (15), with no single model accounting 
for more than 8~cells. These represent question-specific strengths 
rather than systematic capability, consistent with the high 
inter-model correlations ($r > 0.85$) reported in the main text.

\subsection*{Supplementary Note: Domain transfer prediction}
\label{snote:domain_transfer}

Leave-one-domain-out cross-validation tested whether NLP features 
trained on five domains could predict fidelity deviation in the 
held-out sixth domain (random forest, 100 estimators). Cross-validated 
$R^2$ was near zero or negative for most domains (Coding: $-0.04$; 
Conversational: $+0.03$; Domain Knowledge: $-0.19$; Mathematics: 
$+0.05$; Reasoning: $-0.19$; Safety: $+0.26$), confirming that 
NLP features do not generalise across domains for drift prediction.
The Safety domain was the only positive outlier ($R^2 = 0.26$), 
possibly because safety-related responses exhibit more detectable 
surface-level patterns (e.g., refusal language, hedging). Tier 
classification transferred more robustly (accuracy 76--86\% across 
domains), suggesting the binary tier structure is more domain-general 
than the continuous drift values.


\section*{Supplementary Code}

The complete analysis pipeline---including the fidelity deviation 
computation (Python~3.11; NumPy~1.26, SciPy~1.12, pandas~2.1), 
all statistical tests, bootstrap rank analysis, and the linear 
mixed-effects model (R~4.3.2; lme4~\cite{bates2015lme4})---is 
deposited in the study repository at 
the Nature Portfolio repository (DOI to be assigned upon publication) under 
CC~BY~4.0. The pipeline comprises:
\begin{itemize}
    \item \texttt{deviation\_computation.py} --- Per-evaluation, 
        model--question, model-level, and domain-level fidelity 
        deviation aggregation (Equations~1--3).
    \item \texttt{statistical\_analysis.py} --- Two-way ANOVA, 
        Kruskal--Wallis tests, bootstrap rank analysis (10,000 
        resamples), pairwise Cohen's $d$ (45 pairs), TOST 
        equivalence tests (28 ceiling pairs), and inter-model 
        correlation matrices.
    \item \texttt{mixed\_effects\_model.R} --- Linear mixed-effects 
        model (Equation~4), ICC extraction, and 
        Nakagawa--Schielzeth $R^2$ computation.
\end{itemize}
Estimated computation time for full reproduction: $<$30~minutes on a 
standard laptop (Intel Core~i7 or equivalent, 16~GB RAM; no GPU 
required). All analyses used fixed random seeds for exact reproducibility.


\section*{Supplementary Software Versions}

All analyses were conducted using the software versions listed below. 
Version pinning ensures full reproducibility of all reported statistics.

\begin{table}[H]
\centering
\caption{\textbf{Software versions used in all analyses.} Python 
packages were managed via \texttt{pip} with a frozen 
\texttt{requirements.txt}; R packages were installed from CRAN 
snapshots dated February 2026.}
\label{stab:software}
\small
\begin{tabular}{@{}ll|ll@{}}
\toprule
\textbf{Python Package} & \textbf{Version} & \textbf{R Package} & \textbf{Version} \\
\midrule
Python        & 3.11.7  & R          & 4.3.2 \\
NumPy         & 1.26.4  & lme4       & 1.1-35.1 \\
SciPy         & 1.12.0  & psych      & 2.3.12 \\
pandas        & 2.2.0   & boot       & 1.3-28.1 \\
statsmodels   & 0.14.1  & effsize    & 0.8.1 \\
scikit-learn  & 1.4.0   & multcomp   & 1.4-25 \\
openpyxl      & 3.1.2   & MuMIn      & 1.47.5 \\
matplotlib    & 3.8.3   & TOSTER     & 0.8.3 \\
seaborn       & 0.13.2  & irr        & 0.84.1 \\
\bottomrule
\end{tabular}
\end{table}


\section*{Supplementary Repository Contents}

The following materials are deposited at the Nature Portfolio repository (DOI to be assigned upon publication) under CC~BY~4.0 licence. During peer review, all materials listed below are available from the corresponding author upon reasonable request:

\begin{enumerate}
\item \texttt{participant\_ratings.csv} --- Anonymised 29,140-cell 
    participant-level rating matrix ($47 \times 10 \times 62$), 
    containing raw Likert scores and computed fidelity deviation 
    values for every evaluation.
\item \texttt{deviation\_computation.py} --- Executable Python pipeline 
    reproducing all fidelity deviation values from raw ratings, 
    including per-evaluation, model--question, model-level, and 
    domain-level aggregation.
\item \texttt{statistical\_analysis.py} --- Scripts reproducing all 
    reported statistics: two-way ANOVA, Kruskal--Wallis tests, 
    bootstrap rank analysis (10,000 resamples), pairwise Cohen's $d$ 
    (45 pairs), TOST equivalence tests (28 ceiling pairs), 
    inter-model correlation matrices, and variance decomposition.
\item \texttt{mixed\_effects\_model.R} --- R script for the linear 
    mixed-effects model (Equation~4), ICC extraction, 
    and Nakagawa--Schielzeth $R^2$ computation.
\item \texttt{questions\_and\_references.json} --- All 62 prompts 
    with system messages and expert-validated reference answers, 
    organised by domain.
\item \texttt{evaluation\_rubric.pdf} --- Complete 5-point Likert 
    rubric with anchor definitions, calibration protocol, and 
    worked examples used during evaluator training.
\item \texttt{model\_versions.csv} --- Model identifiers, 
    organisations, access method (web UI), access dates 
    (December 2025--February 2026), and context window sizes.
\item \texttt{nlp\_similarity\_scores.csv} --- Automated NLP 
    similarity metrics (semantic similarity, lexical overlap, POS 
    alignment, sentiment agreement, length ratios) for all 620 
    model--question cells.
\item \texttt{} --- High-resolution versions of all main 
    and supplementary figures in PNG format.
\end{enumerate}

\end{document}